\documentclass[twocolumn, switch]{article} %

\usepackage{preprint}

\usepackage{amsmath, amsthm, amssymb, amsfonts}

\usepackage[numbers,square]{natbib}
\usepackage[utf8]{inputenc}	%
\usepackage[T1]{fontenc}	%
\usepackage{xcolor}		%
\usepackage[colorlinks = true,
            linkcolor = blue,
            urlcolor  = magenta,
            citecolor = cyan,
            anchorcolor = black]{hyperref}	%
\usepackage{booktabs} 		%
\usepackage{nicefrac}		%
\usepackage{microtype}		%
\usepackage{lineno}		%
\usepackage{float}			%

\usepackage{lipsum}		%

\usepackage{bm}
\usepackage{cancel}

\usepackage{newfloat}
\DeclareFloatingEnvironment[name={Supplementary Figure}]{suppfigure}
\usepackage{sidecap}
\sidecaptionvpos{figure}{c}

\usepackage{titlesec}
\titlespacing\section{0pt}{12pt plus 3pt minus 3pt}{1pt plus 1pt minus 1pt}
\titlespacing\subsection{0pt}{10pt plus 3pt minus 3pt}{1pt plus 1pt minus 1pt}
\titlespacing\subsubsection{0pt}{8pt plus 3pt minus 3pt}{1pt plus 1pt minus 1pt}

\usepackage{tikz,xcolor,hyperref}

\definecolor{lime}{HTML}{A6CE39}
\DeclareRobustCommand{\orcidicon}{
	\begin{tikzpicture}
	\draw[lime, fill=lime] (0,0)
	circle [radius=0.16]
	node[white] {{\fontfamily{qag}\selectfont \tiny ID}};
	\draw[white, fill=white] (-0.0625,0.095)
	circle [radius=0.007];
	\end{tikzpicture}
	\hspace{-2mm}
}

\foreach \x in {A, ..., Z}{\expandafter\xdef\csname orcid\x\endcsname{\noexpand\href{https://orcid.org/\csname orcidauthor\x\endcsname}
			{\noexpand\orcidicon}}
}
\title{Humanoid Robot Acrobatics Utilizing Complete Articulated Rigid Body Dynamics}

\usepackage[firstpageonly=true,angle=0,scale=0.06,colorspec=0.5,vpos=0.96\paperheight]{draftwatermark}
\usepackage{randtext}
\SetWatermarkText{*correspondence: \randomize{gerald.brantner@gmail.com} }

\usepackage{authblk}

\author[1\thanks{\tt{gerald.brantner@gmail.com}}]{Gerald Brantner}

\affil[1]{Unaffiliated}
\begin{document}

\twocolumn[ %
  \begin{@twocolumnfalse} %

\maketitle

\vspace*{-0.5cm}

\begin{center}
  \url{https://cuboid12.vercel.app}
\end{center}

\vspace*{0.5cm}

\begin{abstract}

  Endowing humanoid robots with the ability to perform highly dynamic motions akin to human-level acrobatics has been a long-standing challenge. 
  Successfully performing these maneuvers requires close consideration of the underlying physics in both trajectory optimization for planning and control during execution. This is particularly challenging due to humanoids' high degree-of-freedom count and associated exponentially scaling complexities, which makes planning on the explicit equations of motion intractable. 
  Typical workarounds include linearization methods and model approximations. However, neither are sufficient because 
  they produce degraded performance on the true robotic system. 
  This paper presents a control architecture comprising trajectory optimization and whole-body control, intermediated by a matching model abstraction, that enables the execution of acrobatic maneuvers, including constraint and posture behaviors, conditioned on the unabbreviated equations of motion of the articulated rigid body model. 
  A review of underlying modeling and control methods is given, followed by implementation details including model abstraction, trajectory optimization and whole-body controller. The system's effectiveness is analyzed in simulation.
\end{abstract}
\vspace{0.35cm}

  \end{@twocolumnfalse} %
] %

\section{Introduction}

Enabling humanoid robots to execute highly dynamic locomotion such as jumping, hopping, flipping, spinning, that is akin to human-level acrobatics, has been an abiding ambition of researchers and robot developers. The motivation for developing this ability is the deployment of humanoid robots in human environments, as 
these highly dynamic motions in many aspects represent a generalization of various simpler maneuvers including walking, running, and balancing. Once humanoid robots can perform motions at the level of human acrobats, nearly all environments that humans operate in become accessible to robots.

Much research effort has been dedicated and much progress has been made towards humanoid walking. 
Two prominent foundational methods in humanoid walking are the ZMP method and their various extensions \cite{vukobratovic2004zero, sardain2004forces, goswami1999postural, spong2003further}, and the SLIP model \cite{blickhan1993similarity}. 
Early work on angular momentum based walking controllers include \cite{goswami2004rate} and \cite{sano1990realization}. 

 A most seminal branch of research are the inherently dynamic methods broached in \cite{raibert1984hopping, raibert1986legged} that evolved into highly capable robotic systems by Boston Dynamics \cite{guizzo2019leaps}. Their implementation technicalities however remain largely unpublished. 
 Recent developments towards jumping and highly dynamic motions include 
 \cite{qi2023vertical}, which utilized centroidal angular momentum control together with an inverse dynamics solver realizing a vertical jump.  
 \cite{dai2014whole} employed a simplified centroidal dynamics model plus full kinematics to generate motion plans for acrobatic motions. 
\cite{he2024cdm} presented a dynamic planning and control framework for jumping motions, approximating legs as constant density cuboid with variable dimensions. 
 \cite{wang2023unified} demonstrated a 90-degree twisting jump and forward jump modeling a single rigid body for the robot's base and point masses for the legs as a centroid. 
\cite{bang2024variable} utilized a pre-trained CNN approximating the robots centroidal inertia tensor and demonstrated fast bipedal walking maneuvers.

Highly dynamic locomotion, as revealed in the term itself, requires the close consideration of the robot's underlying physics, i.e. equations of motion. This presents a particular challenge for humanoid robots due to their high count of degrees-of-freedom  and exponentially scaling complexities and computation times. Computing trajectory optimizations accounting for the equations of motion explicitly is practically intractable, particularly when computation time is essential e.g. in MPC type of architectures.
Whole-body controllers are able to operate in high-dof spaces but are alone insufficient because complex motions require a planning/optimization horizon over an extended period of time. 

Typical workarounds to these challenges have been a) limiting the robot to quasi-static motions e.g. walking where fast dynamics can be neglected, b) relying on model simplifications and approximations that do allow for fast trajectory optimization, but, because they do not accurately reflect the true dynamics, lead to degraded performance on the actual robotic system, or c) constraining the underlying robot design to compositions that better match these model simplifications, e.g. by designing light-weight limbs and heavy torsos.

This paper presents a control architecture comprising trajectory optimization and whole-body control intermediated by a matching model abstraction that allows for planning and control without model approximations of the articulated rigid body model. Essential ingredients of the controller include the control of center-of-mass, angular momentum and its subspaces, inertia-shaping, constraints (joint-limits, collision avoidance), and posture control. 
Utilizing this architecture and complete underlying dynamics enables the execution of highly dynamic, acrobatic motions.

Section \ref{sec_preliminaries} provides a review of the underlying whole-body controller and preliminaries on centroidal dynamics. Section \ref{sec_planning} describes the intermediary model abstraction and trajectory optimization. 
Section \ref{sec_controller} provides the whole-body controller implementation details on launch, flight, and landing sequences. 
Section \ref{sec_simulation} demonstrates the effectiveness and analytics in simulation on various acrobatic maneuvers including leaps, flips and twisting jumps.

\section{Preliminaries}
\label{sec_preliminaries}

\subsection{Operational-Space based Whole-Body Control}
\label{sec_op_preliminaries}

The equations of motion of a robot modeled as articulated rigid body dynamical system can be expressed as 

\begin{equation}
\bm{M}(\bm{q}) \ddot{\bm{q}} + \bm{b}(\bm{q}, \dot{\bm{q}} ) + \bm{g}(\bm{q}) = \bm{\Gamma} + \bm{J_s}^{\top} \bm{F_s}
\end{equation}

where $\bm{q}$ is the the vector of $n$ generalized (minimal) coordinates, containing both active and passive degrees-of-freedom. $\bm{M}$ is the kinetic energy (or mass) matrix, $\bm{b}$ is the vector containing centrifugal and coriolis terms, $\bm{g}$ is the vector of gravitational terms, $\bm{\Gamma}$ is the vector of generalized forces applied by the controller, $\bm{F_s}$ are external support forces (e.g. foot contact) and $\bm{J_s}$ the associated Jacobian.

The operational-space based whole-body controller (hereafter abbreviated to OPWBC) is well-established \cite{khatib2008unified}. Its essential structure and terms are briefly recited here. The dynamic behavior (disregarding contact forces) of a so-called task associated with the jacobian $\bm{J_t}$ can be obtained by 

\begin{equation}
\bar{\bm{J}}_t^\top[\bm{M} \ddot{\bm{q}}+\bm{b}+\bm{g}=\bm{\Gamma}] \Longrightarrow \bm{\Lambda}_t \dot{\bm{\vartheta}_t}+\bm{\mu}_t+\bm{p}_t=\bar{\bm{J}}_t^\top \bm{\Gamma} = \bm{F}_t,
\end{equation}

where $\bar{\bm{J}}_t$ is the dynamically consistent generalized inverse of $\bm{J}_t$ \cite{khatib1990motion} and $\bm{\vartheta}$ is the operational-space coordinate space. $\bm{\Lambda}_t$, $\bm{\mu}_t$, $\bm{p}_t$ are the (pseudo) kinetic energy, coriolis/centrifugal, and gravitational terms associated with the task, respectively. $\bm{F}_t$ is the task control force. 

The task controller can be established as 

\begin{equation}
\bm{F}_t = \hat{\bm{\Lambda}}_t \bm{F}_t^* + \hat{\bm{\mu}}_t + \hat{\bm{p}}_t. 
\end{equation}

Assuming accurate estimates for $\hat{\cdot}$ quantities, linear control laws $\bm{F}_t^*$ can be applied to the resulting decoupled unit-mass system.

The control torque $\bm{\Gamma}$ can be decomposed into 

\begin{equation}
\bm{\Gamma} = \bm{J}_t^\top \bm{F}_T + \bm{N}_t^\top \bm{\Gamma}_p
\end{equation}

where $\bm{\Gamma}_p$ is the control torque associate with posture and $\bm{N}_t$ is the dynamically-consistent null-space projection matrix of the task. 

Establishing decoupled control between task and posture, the posture must be consistent with the task, i.e. controlled in the task's null-space \cite{khatib2004whole}, resulting in the control law

\begin{equation}
\bm{\Gamma} = \bm{J}_t^\top \bm{F}_T + (\bm{J}_p \bm{N}_t)^\top \bm{F}_p.
\end{equation}

The range space of $\bm{J}_p \bm{N}_t$ is the remaining space that's consistent with the task space \cite{khatib2022constraint} that can be computed by SVD decomposition

\begin{equation}
\bm{J}_p \bm{N}_t = \bm{U}_{p|t} \bm{\Sigma}_{p|t} \bm{V}_{p|t},
\end{equation}

leading to the task-consistent posture jacobian 

\begin{equation}
\bm{J}_{p|t} = \bm{U}_{p|t}^\top \bm{J}_p \bm{N}_t
\end{equation}

and its associated control force

\begin{equation}
\bm{F}_{p|t} = \bm{U}_{p|t}^\top \bm{F}_p.
\end{equation}

The decoupled task and posture control law hence becomes 

\begin{equation}
\bm{\Gamma} = \bm{J}_t^\top \bm{F}_t + \bm{J}_{p|t}^\top \bm{F}_{p|t}.
\end{equation}

In a similar fashion, another layer for constraints (e.g. joint limits, (self) collision avoidance) can be constituted:

\begin{equation}
\bm{\Gamma} = \bm{J}_{c}^\top \bm{F}_{c} +  \bm{J}_{t|c}^\top \bm{F}_{t|c} + \bm{J}_{p|t|c}^\top \bm{F}_{p|t|c}.
\end{equation}

Defining the whole-body task representation, corresponding Jacobian and control force
\begin{equation}
\bm{\vartheta}_\otimes = \left[ \bm{\vartheta}_c \ \bm{\vartheta}_{t|c} \  \bm{\vartheta}_{p|t|c} \right] ^\top
\end{equation}
\begin{equation}
\bm{J}_\otimes = \left[ \bm{J}_c \ \bm{J}_{t|c} \  \bm{J}_{p|t|c} \right] ^\top
\end{equation}
\begin{equation}
\bm{F}_\otimes = \left[ \bm{F}_c \ \bm{F}_{t|c} \  \bm{F}_{p|t|c} \right] ^\top,
\end{equation}

the whole-body dynamic equations of motion and control torque become

\begin{equation}
\bm{\Lambda}_\otimes \bm{\dot{\vartheta}_\otimes}+\bm{\mu}_\otimes+\bm{p}_\otimes = \bm{F}_\otimes
\end{equation}

\begin{equation}
\bm{\Gamma} = \bm{J}_\otimes^\top \bm{F}_\otimes,
\end{equation}
respectively.
Taking support in consideration, the dynamically consistent null-space $\bm{N}_s$ and dynamically consistent generalized inverse of the constrained Jacobian $\bar{\bm{J}}_{\otimes|s}$ associated with the support Jacobian $\bm{J}_s$ produces the whole-body dynamic equations of motion under support

\begin{equation}
\bm{\Lambda}_{\otimes|s} \dot{\bm{\vartheta}}_{\otimes|s}+\bm{\mu}_{\otimes|s}+\bm{p}_{\otimes|s} = \bar{\bm{J}}_{\otimes|s}^\top \bm{N}_s^\top \bm{\Gamma}.
\end{equation}

As mentioned above, $\bm{q}$ consists of active and passive coordinates. The passive coordinates comprise the six free-floating coordinates that mobilize the torso. Hence, the system is underactuated and the mapping between actuated ($\bm{\Gamma}_a$) and extended control torques ($\bm{\Gamma}$) is

\begin{equation}
\bm{\Gamma} = \bm{U}^\top \bm{\Gamma}_a,
\end{equation}
where $\bm{U}$ is a selection matrix. This redundancy can be resolved, for instance, with using a dynamically consistent generalized inverse of $\bm{U}\bm{N}_s$.

\subsection{Centroidal quantities}
\label{sec_prelim_centroid}

The centroidal model establishes an
aggregation of a multi-body dynamics model 
, where the entire robot's inertial properties, at a give configuration $\bm{q}_\mathrm{config}$, can be summed up into and made equivalent to a single rigid body with the same inertial quantities. \cite{orin2013centroidal} provides a thorough introduction and computational methods for centroidal quantities.

The centroidal mass or linear inertia is computed by

\begin{equation}
m_\mathrm{centroid} = \sum_{ i \in \mathrm{Bodies}} m_i , 
\end{equation}

where $m_i$ refers to the mass of body $i$. The rotational inertia tensor can be computed by

\begin{equation}
^0\bm{I}_\mathrm{centroid}=\sum_{ i \in \mathrm{Bodies}}  { }_i^0 \! \bm{R} \ {}^i\!\bm{I}_i \ {}_i^0 \! {\bm{R}}^{\top} + m_i \left[\bm{\rho}_i \times\right] \left[\bm{\rho}_i \times\right]^{\top}  \ ,
\end{equation}

where ${}^i\!\bm{I}_i$ refers to the inertia tensor of body $i$, expressed about its center-of-mass; $\bm{\rho}_i$ refers to the distance vector between the centroidal's center-of-mass and body $i$'s center-of-mass, embedded in a skew-symmetric matrix.

The system's linear momentum is 
\begin{equation}
\bm{h}_\mathrm{lin} = m_\mathrm{centroid} \ \bm{v}_\mathrm{com}, 
\end{equation}

where $\bm{v}_\mathrm{com}$ refers to the center-of-mass's linear velocity.

The system's angular momentum is related to generalized velocities $\bm{\dot{q}}$ by the Centroidal Momentum Matrix $\bm{H_\mathrm{ang}}$
\begin{equation}
\bm{h}_\mathrm{ang} = \bm{H_\mathrm{ang}} \ \bm{\dot{q}}.
\label{eq_H} 
\end{equation}

This matrix can be computed by

\begin{equation}
\bm{H}_\mathrm{ang} = \sum_{ i \in \mathrm{bodies}} \left(\left( { }_i^0 \! \bm{R} \ {}^i\!\bm{I}_i \ {}_i^0 \! {\bm{R}}^{\top}\right) {}^0\!\bm{J}_{\omega_i}+m_i\left[\bm{\rho}_i \times\right] {}^0\! \bm{J}_{v_i}\right),
\label{eq_h_ang}
\end{equation}

where ${}^0\!\bm{J}_{\omega_i}$ and ${}^0\! \bm{J}_{v_i}$ refer to the $i^\mathrm{th}$ body angular velocity Jacobian and linear velocity Jacobian of its center-of-mass, respectively.
Similar to \ref{eq_h_ang}, the system's angular momentum equation

\begin{equation}
\bm{h}_\mathrm{ang} = \bm{I}_\mathrm{centroid} \ \bm{\omega}_\mathrm{average}, 
\label{ref_eq_h_ang}
\end{equation}

reveals $\bm{\omega}_\mathrm{average}$, an  \emph{average angular velocity} term as discussed in \cite{essen1993average} (which is noteworthy, however not an essential term for the prescribed control scheme).
The centroidal quantities described above can be and are frequently expressed in spatial representations. But, due to task separation and prioritization (as described below), they are detached here.

\section{Planning and Trajectory Optimization}
\label{sec_planning}

\subsection{Overview}

The execution of highly-dynamic motions requires a) a trajectory optimization with an extended time horizon and b) a controller enforcing the planned trajectory during execution. Both components need awareness of the precise underlying physics i.e. equations of motion because approximated dynamic models inevitably lead to degraded performance and significant deviations that are unacceptable on the real robotic system.
Trajectory optimization utilizing the complete (and highly non-linear) equations of motions of a humanoid robot in generalized coordinate space, however, is practically intractable due to exponentially scaling complexities and computation times. 
A commonly used model simplification is the so-called potato model where the entire robot is reduced to a single rigid body, i.e. a centroid with constant inertial properties.

\begin{figure}[H]
  \centering
  \includegraphics[width=.48\textwidth]{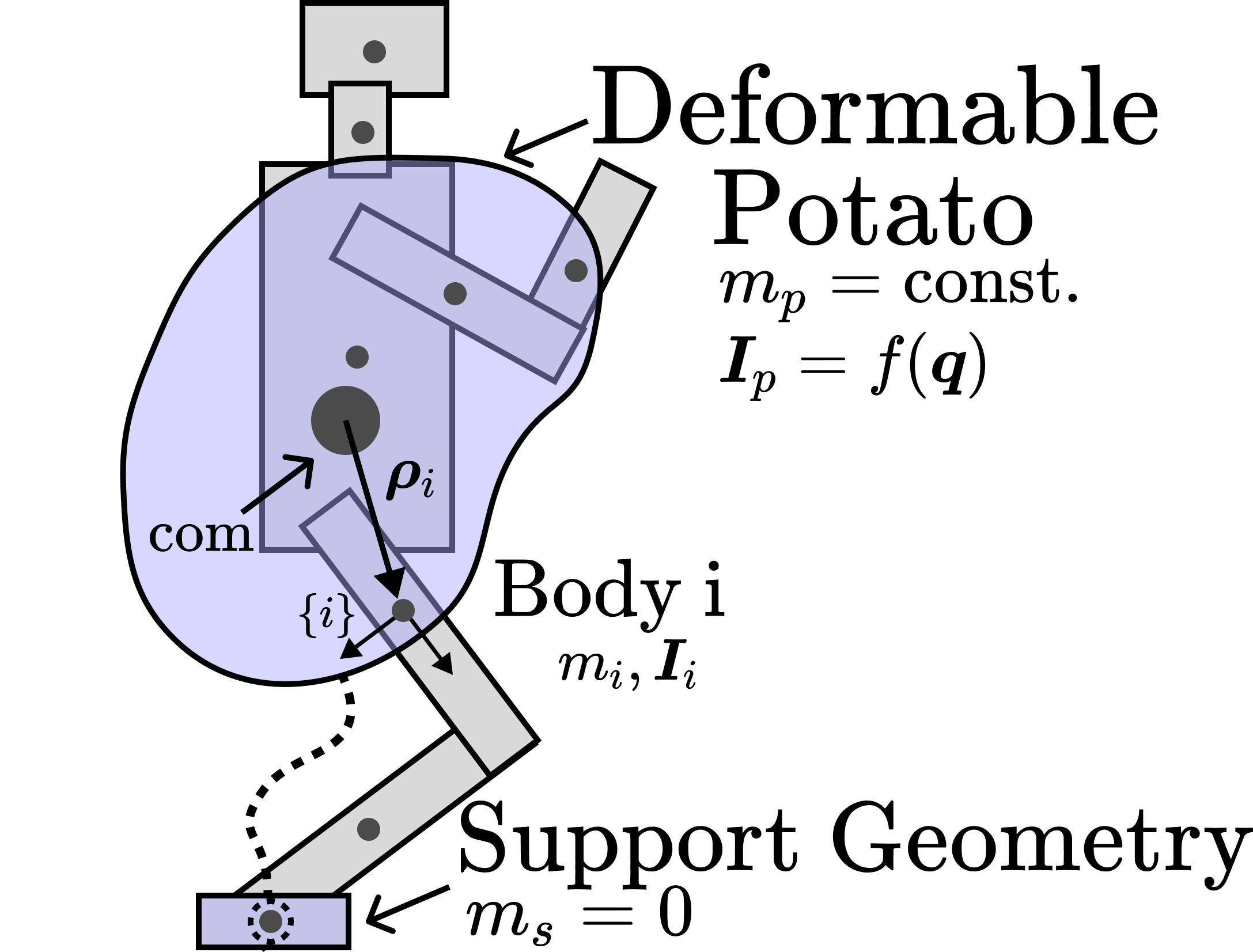}
  \caption{ 
  \textbf{Strangely-supported-deformable-potato (SSDP) model.} The gray components represent the robot's individual bodies, each modeled as rigid bodies. The purple components indicate the model abstraction 'seen' by the trajectory optimizer.
  All the individual bodies inertial properties are reflected into the (larger purple) deformable potato with constant total mass but variable inertia. The (smaller purple) support surface is mass-less and disjointed as the robot's kinematics are hidden.}
  \label{fig_deformable_potato}
\end{figure}

One critical component of the method presented here is the choice of an abstraction that bridges trajectory optimization in low-dimensional spaces allowing fast computation times, with a matching OPWBC controller, where both can rely on the true underlying articulated rigid body dynamics, without approximations. This abstraction can best (however somewhat colloquially) be described as a \textbf{strangely supported deformable potato or SSDP model}  (Figure \ref{fig_deformable_potato}).
\emph{Deformable potato} referring to the above-mentioned potato model, where the robot's entire inertial properties are captured in a single rigid body with constant total mass (i.e. centroid), but with the extension of an adjustable rotational inertia tensor. \emph{Strangely supported}  referring to the addition of support geometries (e.g. a foot) through which external forces can be applied that appear disconnected and mass-less from the perspective of the trajectory optimizer.

\begin{figure}[H]
  \centering
  \includegraphics[width=.48\textwidth]{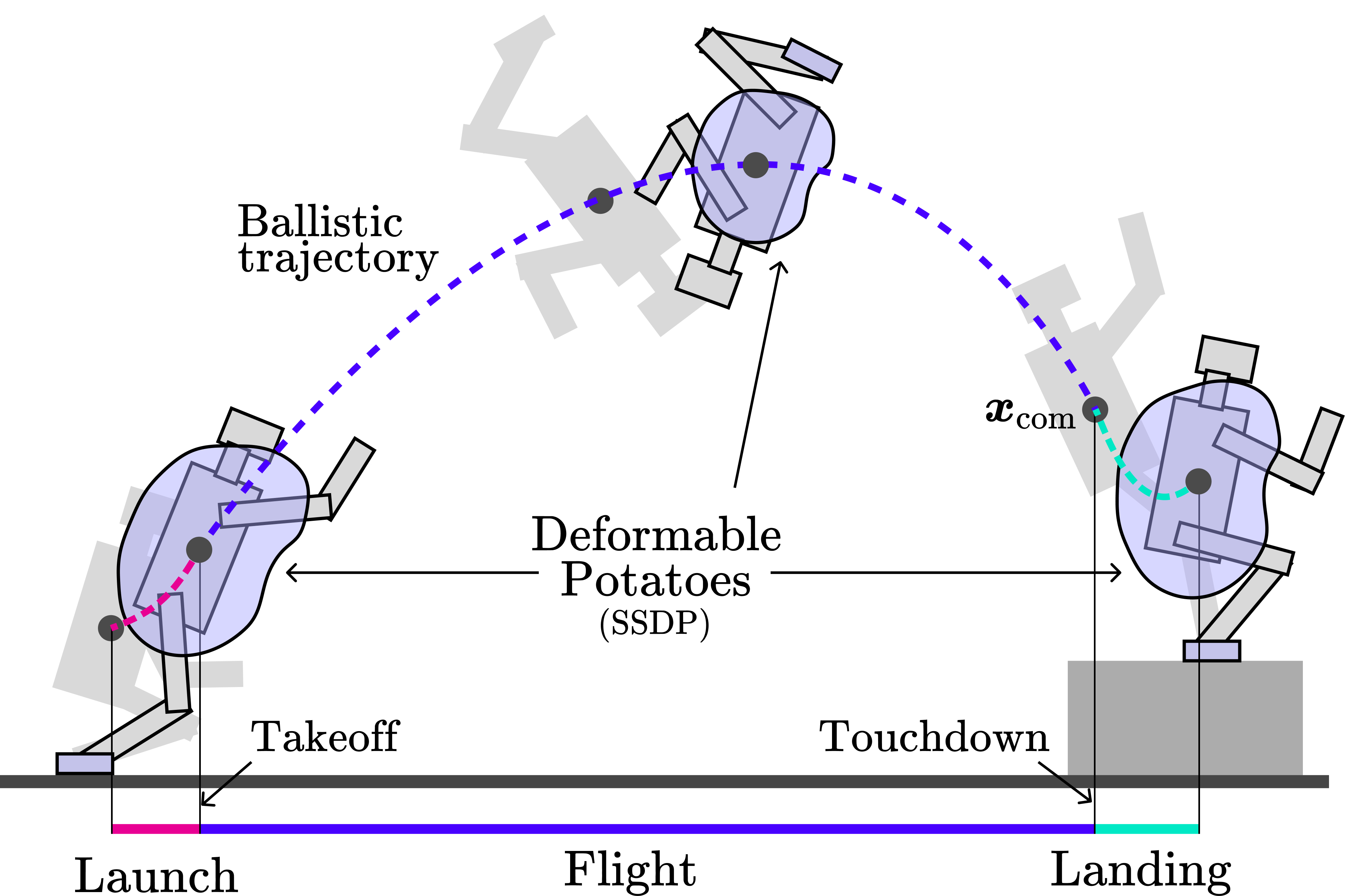}
  \caption{ \textbf{Jump phases.}    Magenta: Launch phase, blue: flight phase, green: landing phase. The robot's center-of-mass follows a ballistic trajectory during flight. Launch and Landing phases include external contact forces, Flight does not.}
  \label{fig_figma_phases}
\end{figure}

A jump (or acrobatic) motion broadly consists of three phases: launch, flight, and landing (Figure \ref{fig_figma_phases}). During launch, the robot initiates at a steady-state configuration with its center-of-mass within the foot support zone and accelerates itself through the application of external ground contact forces to be optimally conditioned for the flight phase.    Much of the subsequent phases is determined by the launch, as during flight no external contact forces can be applied and the robot's center-of-mass follows a ballistic trajectory. During landing, the robot dissipates its kinetic and potential energies, again, through the application of ground contact forces, back into a steady state with its center-of-mass within its support zone.

\subsection{Launch}

\begin{figure}[H]
  \centering
  \includegraphics[width=.48\textwidth]{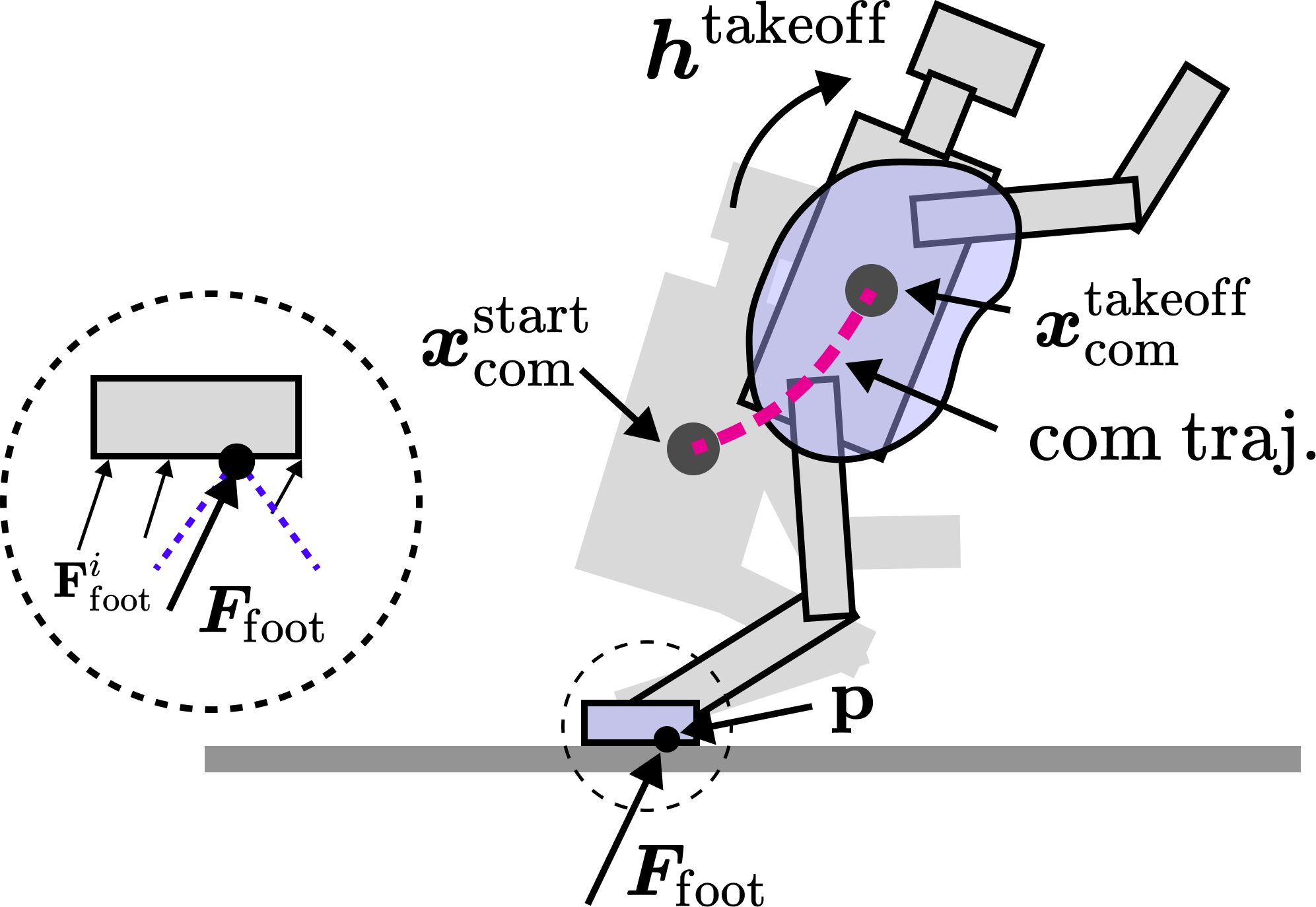}
  \caption{ \textbf{Launch.} Center-of-mass $\bm{x}_\mathrm{com}$ and angular momentum $\bm{h}$ are accelerated to desired takeoff goals. Resultant foot contact force $\bm{F}_\mathrm{foot}$, constrained by static friction limit (purple), applies through the center-of-pressure $\bm{p}$.}
  \label{fig_launch_figma}
\end{figure}

The launch phase is defined as the period from initialization until the robot detaches from the environment. During launch, external contact forces can be exerted on the robot.
The purpose of the launch phase (Figure \ref{fig_launch_figma}) is to accelerate the robot's center-of-mass ($\bm{x}_\mathrm{com}$) and angular momentum ($\bm{h}$) to desired states at the moment of \emph{takeoff} that optimally conditions it to accomplishing a desired flight trajectory. The model assumes a static friction contact between the ground and the foot. Contact forces $\bm{F}_\mathrm{foot}^i$ apply and are combined into a resultant contact force $\bm{F}_\mathrm{foot}$ (analogous extension to contact wrenches).

The problem statement can be transcribed into the following quadratic program (QP). The program assumes $N$ knot point and a constant time interval $dt$. One option for the cost function is the sum square norm of the contact forces. Constraints include the system's discretized dynamics (forward-euler here), contact friction limits, center-of-pressure location within the support zone, and path constraints of the robot's center-of-mass.

\begin{equation}
\begin{aligned}
\underset{\substack{\bm{x}_\mathrm{com}[k],\ \dot{\bm{x}}_\mathrm{com}[k], \\ \ddot{\bm{x}}_\mathrm{com}[k], \  \bm{h}[k], \\  \dot{\bm{h}}[k], \ \bm{F}_\mathrm{foot}[k], \\ \bm{p}[k]  }}{\text{minimize}} \quad & \sum_{k=0}^{N}{ \bm{F}_\mathrm{foot}[k]^\top \bm{F}_\mathrm{foot}[k]   } & \text{(cost)} \\
 \textrm{subject to} \quad &   m \ \ddot{\bm{x}}_\mathrm{com}[k] = \bm{F}_\mathrm{foot}[k] - \bm{g} & \text{(dynamics)}  \\
&  \dot{\bm{x}}_\mathrm{com}[k+1] = \dot{\bm{x}}_\mathrm{com}[k] + dt \ \ddot{\bm{x}}_\mathrm{com}[k] \\
&  \bm{x}_\mathrm{com}[k+1] = \bm{x}_\mathrm{com}[k] + dt \ \dot{\bm{x}}_\mathrm{com}[k] \\
&  \dot{\bm{h}}[k] =  (\bm{p}[k] - \bm{x}_\mathrm{com}[k]) \times \bm{F}_\mathrm{foot}[k]  \\
&  \bm{h}[k+1] = \bm{h}[k] + dt \ \dot{\bm{h}}[k] \\
  & \bm{x}_\mathrm{com}[0] = \bm{x}_\mathrm{com}^\mathrm{start}  & \text{(boundaries)}  \\
  & \dot{\bm{x}}_\mathrm{com}[0] = \dot{\bm{x}}_\mathrm{com}^\mathrm{start}    \\
  & \bm{x}_\mathrm{com}[N] = \bm{x}_\mathrm{com}^\mathrm{takeoff}    \\
  & \dot{\bm{x}}_\mathrm{com}[N] = \dot{\bm{x}}_\mathrm{com}^\mathrm{takeoff}    \\
  & \bm{h}[0] = \bm{h}^\mathrm{start}   \\
  & \bm{h}[N] = \bm{h}^\mathrm{takeoff}    \\
&  \bm{P}  \bm{F}_\mathrm{foot}[k]  \preccurlyeq \bm{0}   & \text{(friction)}  \\
& \bm{A}^\mathrm{support}   \bm{p}[k]  \preccurlyeq \bm{b}^\mathrm{support}   & \text{(cop)}  \\
& \bm{x}_\mathrm{com}^\mathrm{min}  \preccurlyeq   \bm{x}_\mathrm{com}[k]  \preccurlyeq \bm{x}_\mathrm{com}^\mathrm{max}   & \text{(path)}  \\
\end{aligned}
\end{equation}

Notably, this program does not include the robot's generalized coordinates $\bm{q}$, but only the quantities abstracted by the SSDP model, critically reducing the optimization problem's dimensionality. The desired values $\bm{x}_\mathrm{com}^\mathrm{takeoff}$ and $\bm{h}^\mathrm{takeoff}$ can be inferred from the planned flight trajectory, as elaborated in the following section.

 The robot COM's desired take-off state $\bm{x^\mathrm{takeoff}}$ can be inferred from a ballistic target solution. There are various solutions for escape-velocity (or linear momentum) vectors that accomplish a given target position. One option that allows a closed-form solution is to pre-define the launch-angle $\alpha_0$ by  heuristics. In this case, the take-off velocity (magnitude) is computed by
\begin{equation}
    v_0=\sqrt{\frac{d_h^2 g}{d_h \sin( 2 \alpha_0)-2 d_v \cos^2( \alpha_0)}}, 
\end{equation}

where $d_h$ and $d_v$ refer to horizontal and vertical distance, respectively. The escape-velocity vector $\bm{v}_0$ follows by trigonometry. 

A rigid body's (or centroid's) angular momentum at take-off determines its rotation during flight ($h = I \omega$) and final orientation at touch-down ($\beta^\mathrm{touchdown} = \beta^\mathrm{takeoff} + t^\mathrm{flight} \cdot \omega $), hence accomplishing the robot's angular momentum at take-off is vital.

The flight-time is computed by
\begin{equation}
t^\mathrm{flight} = \frac{v_0 \sin(\alpha_0) + \sqrt{(v_0 \sin \alpha_0)^2+2 g d_v}}{g}.
\label{eq_flight_time}
\end{equation}

\subsection{Flight}
\label{sec_planning_flight}

The flight phase begins at the moment of takeoff and terminates at the moment of touchdown, hence, no external forces act on the robot during flight. It is well-understood that a rigid body devoid of external forces conserves linear and angular momentum, and, under the influence of gravity follows a ballistic trajectory. Similarly, by the centroidal equivalency, a multi-body system's COM similarly follows a ballistic trajectory. The system's aggregate angular momentum too is preserved throughout:

\begin{equation}
\bm{h}^\mathrm{takeoff} =  \bm{h}^\mathrm{flight} = \bm{h}^\mathrm{touchdown} = \mathrm{const}.
\end{equation}

The flight phase is essential to compensating control errors present at the take-off instant. The COM's trajectory cannot be influenced once the robot detaches from the ground at takeoff. Favorably, the robot's rate of rotation can be controlled during flight. As indicated in \ref{ref_eq_h_ang}, the total angular momentum is proportional to the system's angular velocities and the system's inertias. Hence, by shaping the robot's inertia, the robot's angular velocities can be controlled indirectly. 

As discussed in (\ref{eq_H}), the angular momentum matrix relates angular momentum and generalized coordinate velocities:

\begin{equation}
\bm{h} = \bm{H} (\bm{q}) \ \bm{\dot{q}}.
\label{hHq}
\end{equation}

When representing the torso's mobilization as 6-dimensional minimal coordinates, (\ref{hHq}) can be decomposed into

\begin{equation}
\bm{h} = 
\bm{h}_\mathrm{lin} + \bm{h}_\mathrm{ang} + \bm{h}_\mathrm{act} = 
\left[ \bm{0}_{3 \times 3} \ \bm{H}_\mathrm{ang} \ \bm{H}_\mathrm{act} \right]
\left[\begin{array}{c}
  \bm{\dot{q}}_\mathrm{lin}  \\
  \bm{\dot{q}}_\mathrm{ang}  \\
  \bm{\dot{q}}_\mathrm{act}  \\
\end{array}\right].
\label{eq_h_ang_components}
\end{equation}

The linear cartesian velocities $\bm{\dot{q}}_\mathrm{lin}$ naturally do not contribute; $\bm{H}_\mathrm{ang}$ is of shape $3 \times 3$ and represents the contribution associated with the robot's angular free-floating velocities; and $\bm{H}_\mathrm{act}$ represents the contribution of the actuated joints' velocities $\bm{\dot{q}}_\mathrm{act}$. 
Further, $\bm{H}_\mathrm{ang}$ can be mapped to the robot's rotational centroidal inertia:

\begin{equation}
\bm{H}_\mathrm{ang} \rightarrow \bm{I}_\mathrm{centroid}.
\label{eq_h_euler_I}
\end{equation}

Presuming vanishing contributions of the active coordinates $\bm{h}_\mathrm{act}$, the system's angular velocities can be directly related to its centroidal inertia by

\begin{equation}
\bm{h} = \bm{h}_\mathrm{ang} + \cancel{ \bm{h}_\mathrm{act} } = \bm{H}_\mathrm{ang} \ \bm{\dot{q}}_\mathrm{ang} = \bm{I}_\mathrm{centroid} \ \bm{\omega}_\mathrm{} = \mathrm{const}.
\end{equation}

Fortunately, $\bm{h}_\mathrm{act}$ can in fact be eliminated by the whole-body controller (see next section), which is critical to utilizing the deformable centroid model for flight trajectory optimization. The robot's orientation at the moment of touchdown results from the accumulation of angular velocities over the flight time:

\begin{equation}
\bm{\zeta}^\mathrm{touchdown} =  
\bm{\zeta}^\mathrm{takeoff} 
+  \int_{0}^{t^\mathrm{flight}} \dot{\bm{\zeta}}  dt,
\label{eq_q_integral}
\end{equation} 

where $\bm{\zeta}$ is an orientation and $\dot{\bm{\zeta}}$ an angular velocity representation equivalent to $\bm{\omega}$. The integral above represents an angular velocity integration, for which various methods depending on orientation representation are available \cite{boyle2017integration}. 
The trajectory optimization program for the flight-phase is:

\begin{equation}
\begin{aligned}
\underset{\substack{\bm{\zeta}[k],\ \dot{\bm{\zeta}}[k],\ \ddot{\bm{\zeta}}[k], \\  \bm{I}[k],\  \dot{\bm{I}}[k],\  \ddot{\bm{I}}[k]  }}{\text{minimize}} \quad 
& \sum_{k=0}^{N}{||  \ddot{\bm{I}}[k]  ||^2_2     } & \text{(cost function)}  \\
 \textrm{subject to} \quad 
&  \dot{\bm{\zeta}}[k+1] = \dot{\bm{\zeta}}[k] + dt \ \ddot{\bm{\zeta}}[k] & \text{(kinematic)} \\
&  \bm{\zeta}[k+1] = \bm{\zeta}[k] + dt \ \dot{\bm{\zeta}}[k] \\
&  \dot{\bm{I}}[k+1] = \dot{\bm{I}}[k] + dt \ \ddot{\bm{I}}[k]  \\
&  \bm{I}[k+1] = \bm{I}[k] + dt \ \dot{\bm{I}}[k] \\
  & \bm{\zeta}[0] = \bm{\zeta}^\mathrm{start}  & \text{(boundaries)}  \\
    & \bm{\zeta}[N] = \bm{\zeta}^\mathrm{touchdown}  \\
  & \dot{\bm{\zeta}}[0] = \dot{\bm{\zeta}}^\mathrm{start}    \\
  & \dot{\bm{\zeta}}[N] = \dot{\bm{\zeta}}^\mathrm{touchdown}    \\
 & \bm{\zeta}^\mathrm{touchdown} = \bm{\zeta}^\mathrm{start} + dt \sum_{k=0}^{N}{ \dot{\bm{\zeta}}[k] }  & \text{(integral cond.)}  \\
  & \bm{h} = \bm{I}[k] \ \dot{\bm{\zeta}}[k]  & \text{(h conservation)}  \\
& \bm{I}^\mathrm{min}  \preccurlyeq   \bm{I}[k]  \preccurlyeq \bm{I}^\mathrm{max}   & \text{('path' constraints)}  \\
\end{aligned}
\end{equation}

The above kinematic constraints and integral conditions are dependent on the orientation representation and must be transcribed accordingly. The operators in these equations are placeholders for these.

\subsection{Landing}

The landing phase starts at the moment of touch down. The purpose is to absorb the robot's kinetic and potential energies (i.e. the centroid's center-of-mass velocity and angular momentum) while observing contact constraints. The landing trajectory optimization problem is similar to the launch problem illustrated above. 

It is also worth pointing out that consecutive landing and launch programs can be integrated into a single program enabling cyclical motions such as running or maneuvering more complex terrains.

\section{Controller}
\label{sec_controller}

\subsection{Overview}
The model underlying the controller is a rigid multi-body dynamical system with a free-floating base. The robot's joints are represented by generalized coordinates $\bm{q}^\mathrm{joints}$. The free-floating base mobilization is represented by 6-dimensional minimal coordinates $\bm{q}^\mathrm{free}$, where the rotational components are represented in euler-angles, which is needed in encoding paths such as flips beyond 360-degree rotations.
The whole-body controller utilizes a set of so-called tasks, each instantaneously accomplishing a control objective. Each task utilizes a distinct dynamics model abstraction appropriate for the task. Importantly, these models are not a simplification or reduction of the complete model but rather projections into lower-dimensional task-spaces that are consistent with the complete model.

Matching the trajectory optimization phases, the controller comprises three major sequential parts: Launch, Flight, and Landing, which are described below.

\subsection{Launch}
\label{sec_launch_controller}

The dynamics associated with the whole-body controller during the launch phase can be described by

\begin{equation}
\bm{\Lambda}_{\otimes|s}^{launch} \ \dot{\bm{\vartheta}}_{\otimes|s}^{launch}+\bm{\mu}_{\otimes|s}^{launch}+\bm{p}_{\otimes|s}^{launch} = \bm{F}_{\otimes|s}^{launch}.
\end{equation}

The control law computing torques to the generalized coordinates is

\begin{equation}
\bm{\Gamma}^{launch} = \bm{J}_{\otimes|s}^{^{launch}\top} \ \bm{F}_{\otimes|s}^{launch}.
\end{equation}

As suggested in section \ref{sec_op_preliminaries}, the whole-body controller during launch consists of three components, task, constraints, and posture.

\begin{equation}
\bm{\vartheta}_\otimes = \left[ \bm{\vartheta}_c \ \bm{\vartheta}_{t|c} \  \bm{\vartheta}_{p|t|c} \right] ^\top
\end{equation}

The task's purpose is controlling the robot's center-of-mass, and its angular momentum. 

\begin{equation}
\bm{\vartheta}_t^{launch} = \left[ \bm{\vartheta}_{com} \ \bm{\vartheta}_{h} \right] ^\top
\end{equation}

where the center-of-mass component is
\begin{equation}
\bm{\vartheta}_{com} = \bm{J}_{com} \ \dot{\bm{q}}
\end{equation}

\begin{equation}
\bm{J}_\mathrm{com} = \frac{1}{m_\mathrm{total}} \sum_{ i \in \mathrm{bodies}} m_i \ \bm{J}_{v_i}
\end{equation}

and the angular momentum component is

\begin{equation}
\bm{\vartheta}_{h} = \bm{J}_{h} \ \dot{\bm{q}}
\end{equation}

where $ \bm{J}_{h}$ is computed as prescribed in section \ref{sec_prelim_centroid}.

\begin{figure}[H]
  \centering
  \includegraphics[width=.48\textwidth]{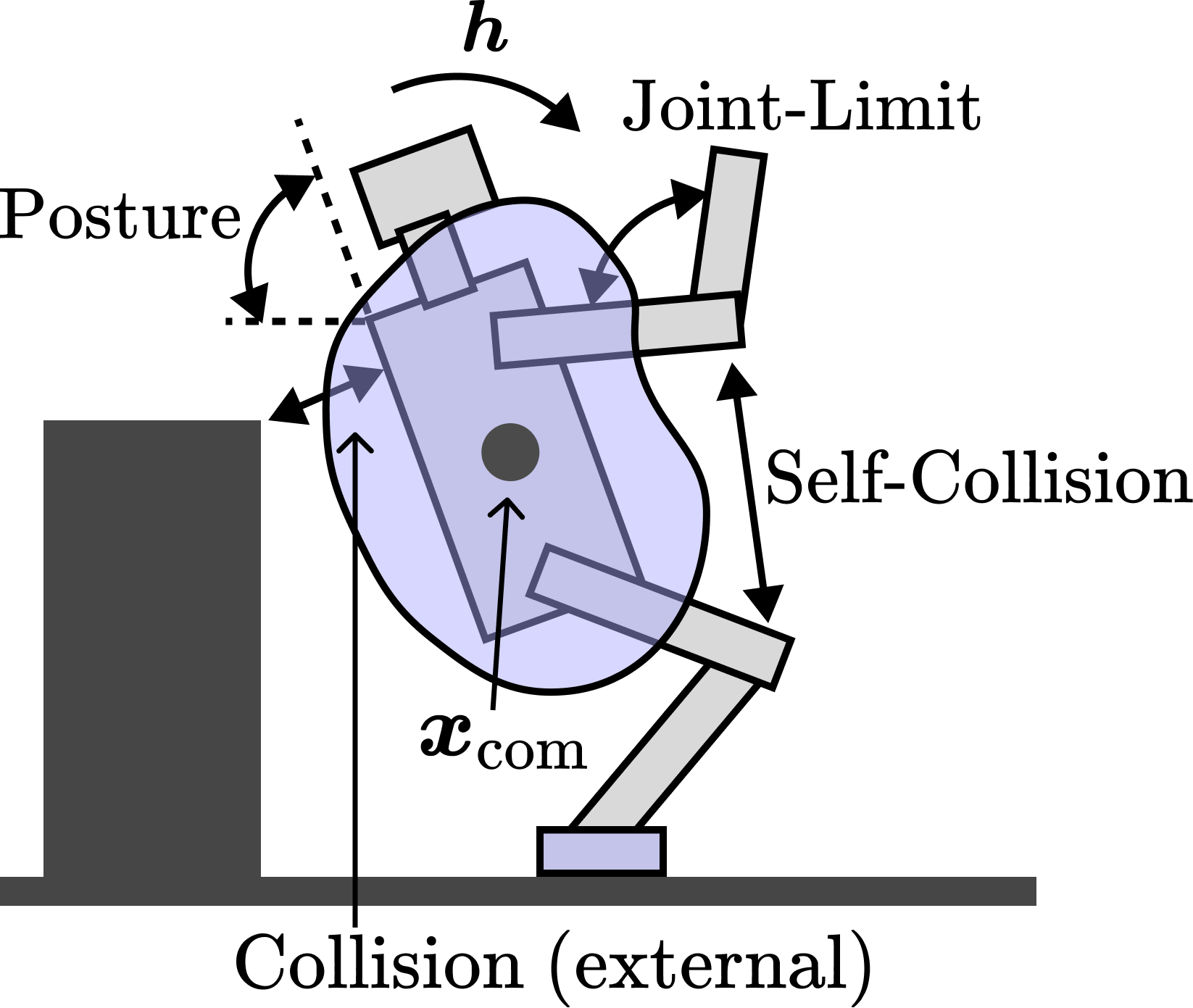}
  \caption{ \textbf{Control Parameters.}  Task: center-of-mass $\bm{x}_{com}$ and angular momentum $\bm{h}$. Constraints: Joint-limit, (external) Collision and Self-Collision avoidance. Posture: Torso orientation (e.g.).}
  \label{fig_figma_control_launch}
\end{figure}

Because the trajectory optimization space is unaware of the robot's generalized coordinates (i.e. joint states), it cannot account for constraints. Hence it is the controller's responsibility to handle these. The constraint superseding the task comprises the avoidance of a) joint-limits (position constraints of the joint actuators), b) external collisions (any external collision between the robot with the environment, other than desired contacts), and c) self-collisions (contact between pairs of non-consecutive links), see Figure \ref{fig_figma_control_launch}. 

\begin{equation}
\bm{\vartheta}_c^\mathrm{launch} = \left[ \bm{\vartheta}_\mathrm{joint\_limit} \ \bm{\vartheta}_\mathrm{collision} \ \bm{\vartheta}_\mathrm{self\_collision} \right] ^\top,
\end{equation}

where

\begin{equation}
\bm{\vartheta}_\mathrm{joint\_limit} = \bm{J}_\mathrm{joint\_limit} \ \dot{\bm{q}}
\end{equation}

\begin{equation}
\bm{\vartheta}_\mathrm{collision} = \bm{J}_\mathrm{collision} \ \dot{\bm{q}}
\end{equation}

\begin{equation}
\bm{\vartheta}_\mathrm{self\_collision} = \bm{J}_\mathrm{self\_collision} \ \dot{\bm{q}}.
\end{equation}

The joint-limit jacobian $\bm{J}_\mathrm{joint\_limit}$ is a selection matrix indicating joints within the so-called activation region. The Jacobians $\bm{J}_\mathrm{collision}$ and $\bm{J}_\mathrm{self\_collision}$ map generalized velocities $\dot{\bm{q}}$ to velocities between the respective collision pairs. 

The so-called Firas function \cite{khatib1986real} is a well-established choice in computing the repulsive forces: 

\begin{equation}
F^*_c =
\begin{cases} 
   -\eta \left( \frac{1}{\rho} - \frac{1}{\rho_0}  \right) \frac{1}{\rho^2} & \text{if} \ \rho \leq \rho \\
   0        & \text{otherwise}
\end{cases},
\end{equation}

where $\eta$ is a constant gain, $\rho$ is a distance metric and $\rho_0$ represents the distance at which the constraint becomes active. Since distance metrics are calculated within the control loop, fast computation times are essential. The distance metric for joint limit control is obvious. Good choices for volumetric representations allowing efficient distance computations for (self-) collision are capsules, planes, and axis-aligned bounding boxes \cite{ericson2004real}. 

The remaining null-space beyond task and constraints can be allocated to the control the robot's posture. Appropriate choices include torso orientation and limb posture (i.e. preferred elbow flare and hip rotation):

\begin{equation}
\bm{\vartheta}_p^\mathrm{launch} = \left[ \bm{\vartheta}_\mathrm{torso\_orientation} \ \bm{\vartheta}_\mathrm{limb\_posture} \right] ^\top.
\end{equation}

Having defined all Jacobians for task, constraints, and posture, the unified Jacobian $\bm{J}_{\otimes}^{^{launch}}$ can be computed according to \ref{sec_op_preliminaries}. 
The support Jacobian $\bm{J}_s$, depending on the contact state, leads to the dynamically-consistent support null-space matrix $\bm{N}_s$

\begin{equation}
\bm{\Lambda}_\mathrm{s}^{-1} = \bm{J}_\mathrm{s} \bm{M}^{-1} \bm{J}_\mathrm{s}^{\top}
\end{equation}

\begin{equation}
\overline{\bm{J}}_s = \bm{M}^{-1} \bm{J}_\mathrm{s}^{\top}   \bm{\Lambda}_\mathrm{s}
\end{equation}

\begin{equation}
\bm{N}_\mathrm{s} = \mathbb{I} - \overline{\bm{J}}_s \bm{J}_\mathrm{s}.
\end{equation}

From here, the unified Jacobian consistent with support is computed by 

\begin{equation}
\bm{J}_{\otimes|s} = \bm{J}_{\otimes} \bm{N}_s,
\end{equation}

and the control force in augmented, support consistent space $\bm{F}_{\otimes|s}^{launch}$ can be computes as prescribed in \ref{sec_op_preliminaries}. Finally the control torque $\bm{\Gamma}^{launch}$ must be mapped to torques of the subset of actuated coordinates by resolving

\begin{equation}
\bm{\Gamma} = \bm{U}^\top \bm{\Gamma}_a.
\end{equation}

\subsection{Flight}

The dynamics associated with the whole-body controller during the flight phase are

\begin{equation}
\bm{\Lambda}_{\otimes}^{flight} \ \dot{\bm{\vartheta}}_{\otimes}^{flight}+\bm{\mu}_{\otimes}^{flight}+\bm{p}_{\otimes}^{flight} = \bm{F}_{\otimes}^{flight}
\end{equation}

and the control law computing torques in generalized coordinates space is

\begin{equation}
\bm{\Gamma}^{flight} = \bm{J}_{\otimes}^{^{flight}\top} \ \bm{F}_{\otimes}^{flight}.
\end{equation}

The robot's center-of-mass cannot be controlled during flight and simply follows the ballistic trajectory determined at takeoff. The main purpose of the flight controller is to ensure the robot touches down at a desired torso orientation.

The flight controller too consists of task, constraint and posture. As mentioned in section \ref{sec_planning}, the robot's rate of rotation can be influenced indirectly by controlling the robot's (centroidal) inertia. This is referred-to as inertia-shaping. Critically, in order for the altered inertia to translate exclusively into torso (i.e. base) coordinates, the contributions to angular momentum of the actuated coordinates $\bm{h}_\mathrm{act}$ must be controlled as well. 

During flight, the robot also needs to condition its foot on which it intends to land at the moment of touchdown. Controlling the foot has equal priority and therefore is part of the task.

The task description hence is

\begin{equation}
\bm{\vartheta}_t^\mathrm{flight} = \left[ \bm{\vartheta}_{I} \ \bm{\vartheta}_\mathrm{h\_act} \ \bm{\vartheta}_\mathrm{foot} \right] ^\top,
\end{equation}

where 

\begin{equation}
\bm{\vartheta}_{I} = \bm{J}_{I} \ \dot{\bm{q}}
\end{equation}

\begin{equation}
\bm{\vartheta}_\mathrm{h\_act} = \bm{J}_\mathrm{h\_act} \ \dot{\bm{q}}
\end{equation}

\begin{equation}
\bm{\vartheta}_\mathrm{foot} = \bm{J}_\mathrm{foot} \ \dot{\bm{q}}.
\end{equation}

The inertia jacobian $\bm{J}_{I}$ is computed by

\begin{equation}
J_I = \frac{\partial \bm{I}}{\partial \bm{q}}.
\end{equation}

As indicated in section \ref{sec_planning_flight}, it is essential to also control the off-diagonal terms of $\bm{I}$, otherwise significant undesired rotations accumulate due to these coupling terms.

The jacobian associated with the actuated joints' contributions to angular momentum $\bm{J}_\mathrm{h\_act}$, as described in section \ref{sec_prelim_centroid}, is

\begin{equation}
J_{\bm{h}_\mathrm{act}} = \frac{\partial \bm{h}_\mathrm{act}}{\partial \bm{q}}.
\end{equation}

The operational space of the foot component comprises both linear position and orientation:

\begin{equation}
\bm{J}_\mathrm{foot} \triangleq \left[\begin{array}{c}
\bm{J}_{v, \mathrm{foot }} \\
\bm{J}_{\omega, \mathrm{foot }} \\
\end{array}\right].
\end{equation}

The constraints component during flight, similarly to above, ensures that joint-limits and collisions are avoided:

\begin{equation}
\bm{\vartheta}_c^\mathrm{flight} = \left[ \bm{\vartheta}_\mathrm{joint\_limit} \ \bm{\vartheta}_\mathrm{collision} \ \bm{\vartheta}_\mathrm{self\_collision} \right] ^\top.
\end{equation}

The posture during flight controls the robot's remaining null-space after task and constraints:

\begin{equation}
\bm{\vartheta}_p^\mathrm{flight} = \left[  \bm{\vartheta}_\mathrm{limb\_posture} \right] ^\top.
\end{equation}

During flight there is no support contact to the environment, therefore no support Jacobians or null-spaces are included.

\subsection{Landing}

The landing controller is similar to the launch controller described in section \ref{sec_launch_controller}. In many ways, landing is the inverse operation of launching. The purpose is the dissipation of kinetic and potential energies while ensuring no violations of contact constraints. 
The controller, again, enforces the optimal trajectories. 

\section{Simulation }
\label{sec_simulation}

The simulations are implemented in the Mujoco \cite{todorov2012mujoco} environment version 3.2 with an integration time step of 1ms and \emph{Fast implicit-in-velocity} integration. The robot is modeled with 18 joints plus free-floating mobilization totaling 24 degrees-of-freedom in simulation. All of the robot's joints are modeled as hinge joints with pure torque actuators, without additional friction or damping. Each link is modeled as a rigid body (cuboid-type geometries), each with associated linear and rotational inertias (except the torso is modeled by three rigidly connected cuboids). IPOPT \cite{wachter2006implementation} version 3.14 is the solver underlying trajectory optimization methods.

The following sections illustrate and analyze a number of maneuvers. In order to avoid redundancy, each simulation is utilized to highlight a particular aspect of the controller.

\subsection{Box Jump}

In this experiment, the robot executes a box-jump manuever, meaning a jump from the ground to a platform at higher elevation. The objective here is to a) execute an appropriate launch trajectory that b) lands the robot at a desired com-position and torso orientation and c) execute a landing trajectory that absorbs the kinetic and potential energies terminating in a stable steady-state. The entire sequence including launch, flight, and landing are illustrated below. 

Figure \ref{fig_grid_boxjump_up} displays renderings of the simulation environment. The robot initializes while balancing on one leg, the goal is to jump and land on the blue box. 
\begin{figure}[H]
  \centering
  \includegraphics[width=.48\textwidth]{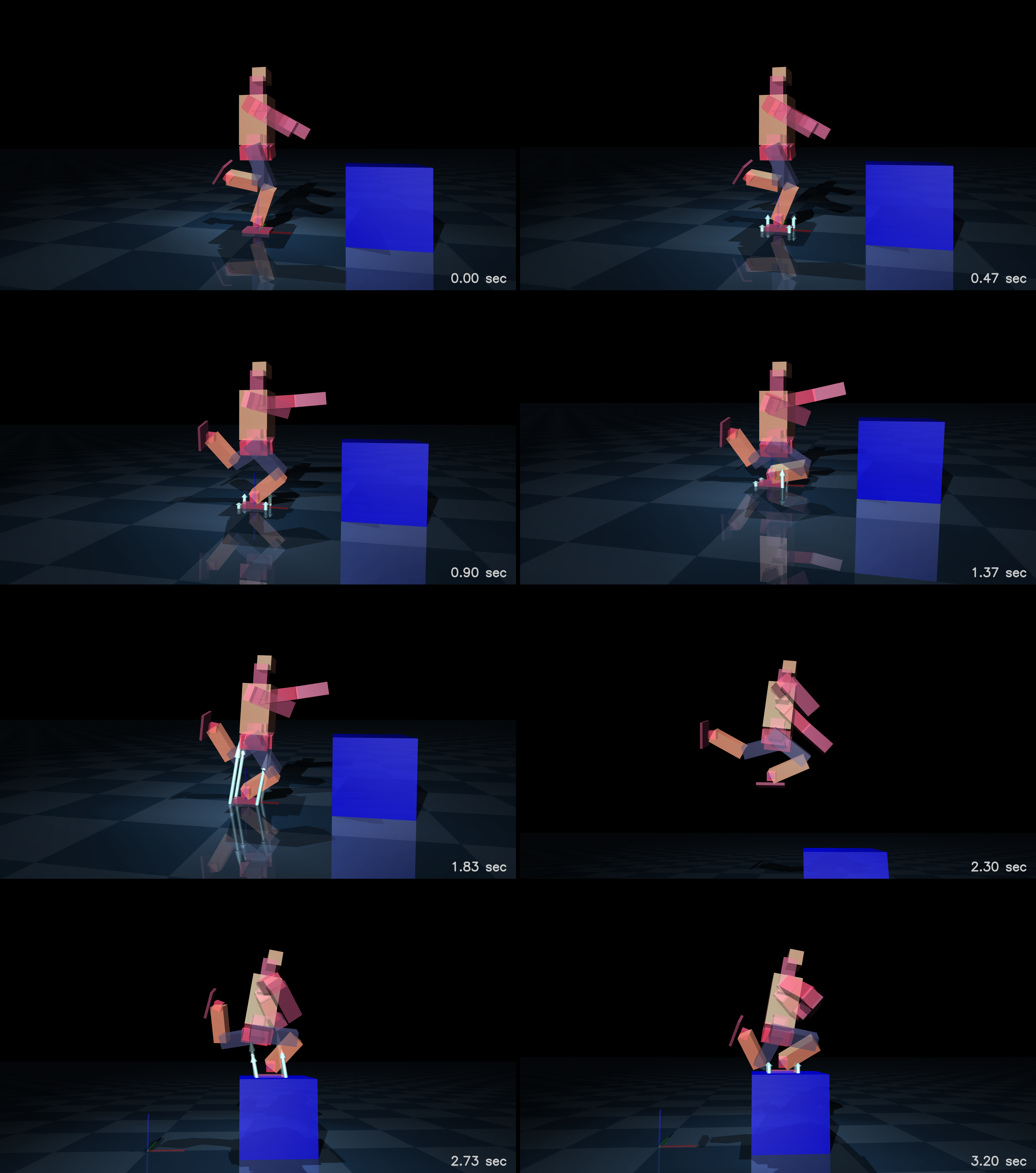}
  \caption{ \textbf{Box Jump Up Rendering.} 1st row: coiling phase, 2nd row: launch phase, 3rd row: flight phase, 4th row: landing phase. The 4 vectors indicate contact forces between foot and ground.}
  \label{fig_grid_boxjump_up}
\end{figure}

\begin{figure}[H]
  \centering
  \includegraphics[width=.48\textwidth]{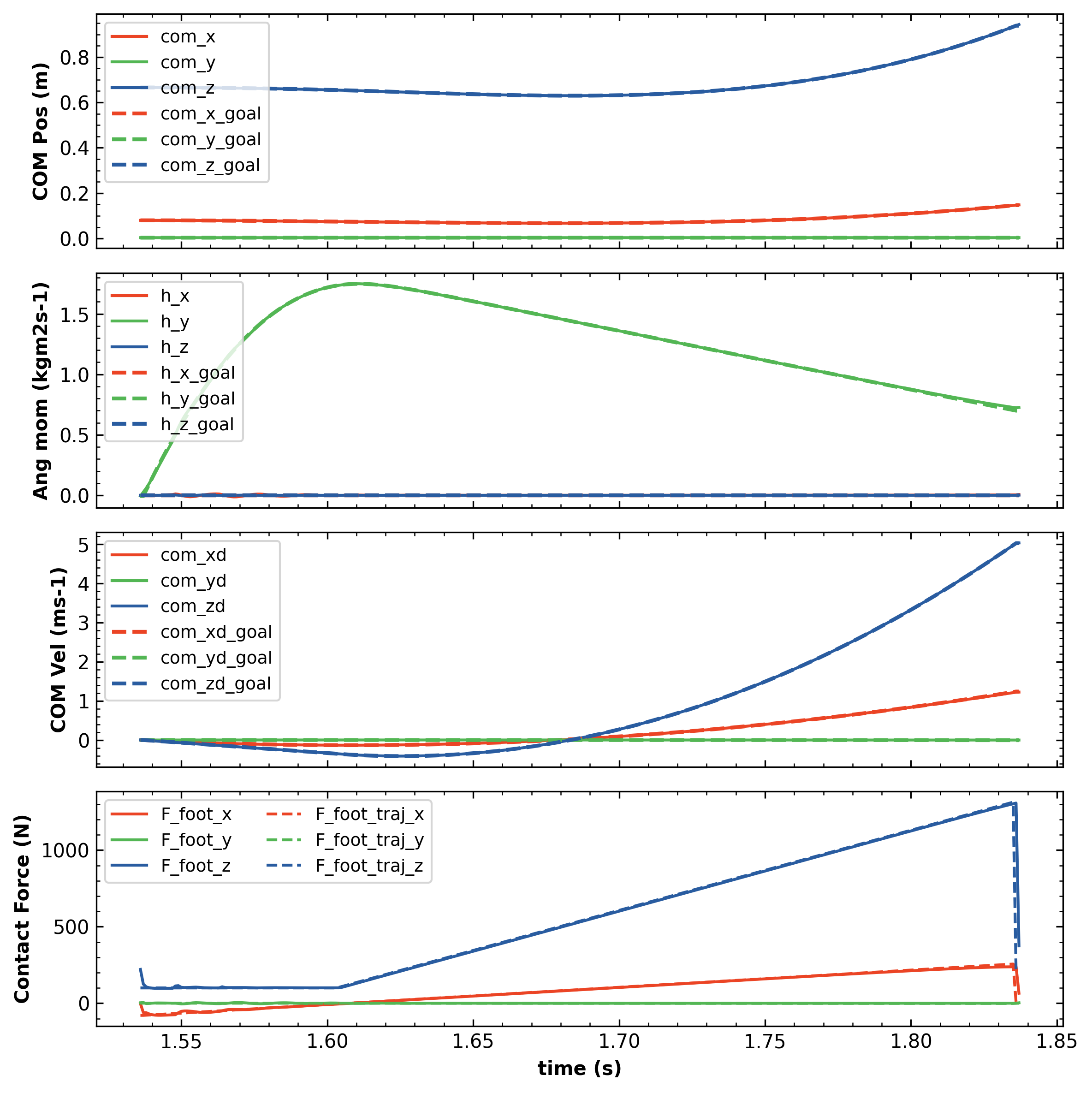}
  \caption{  \textbf{Launch Trajectory Tracking.} Top: center-of-mass, 2nd row: angular momentum, 3rd row: center-of-mass velocity, bottom row: foot-ground resultant contact forces.}
  \label{fig_lineplot_boxjump_up_launch}
\end{figure}

\begin{figure}[H]
  \centering
  \includegraphics[width=.48\textwidth]{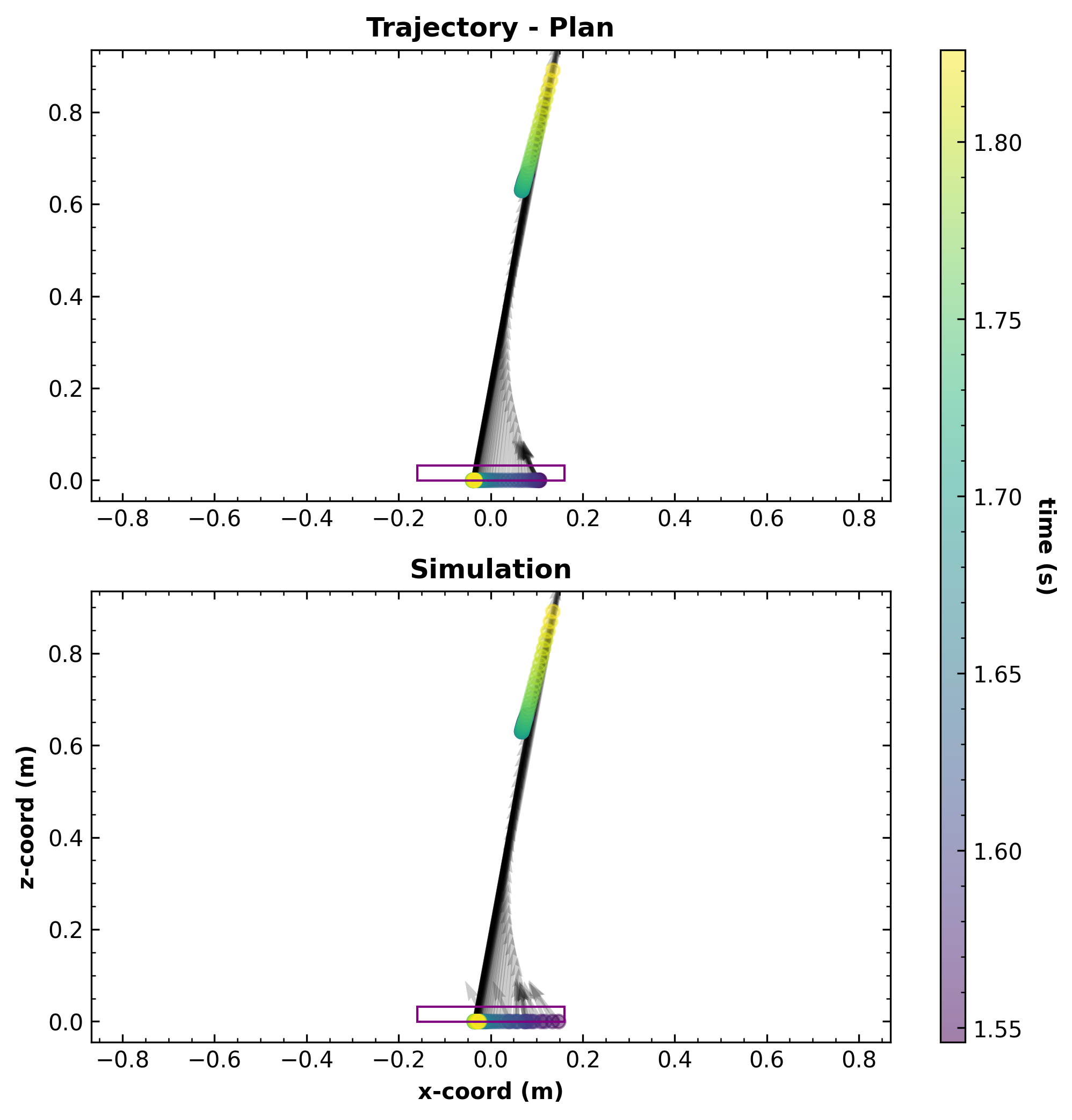}
  \caption{ \textbf{Launch  Scatter Plot.} Top: Planned trajectories. Bottom: Measured simulation states. Black vectors represent resultant ground contact force. The colored point sequences represent center-of-mass (upper) and center-of-pressure (lower). Purple rectangle indicates the foot support. (Plotting every 5th data point.)}
  \label{fig_quiver_boxjump_up_launch}
\end{figure}

Figures \ref{fig_lineplot_boxjump_up_launch} and \ref{fig_quiver_boxjump_up_launch}  quantitatively illustrate trajectory tracking during the launch phase. The purpose of this phase is a) accelerating the robot's center-of-mass to a predetermined take-off velocity at a final position while b) maintaining balance (contact forces remaining within the foot support) and avoiding slippage (contact forces confined within the static friction boundaries).

The controller tracks the desired trajectories with minor deviations and the robot successfully executes the desired maneuver, jumping and landing on the box within the constraints while maintaining stability at all times. 

\begin{figure}[H]
  \centering
  \includegraphics[width=.48\textwidth]{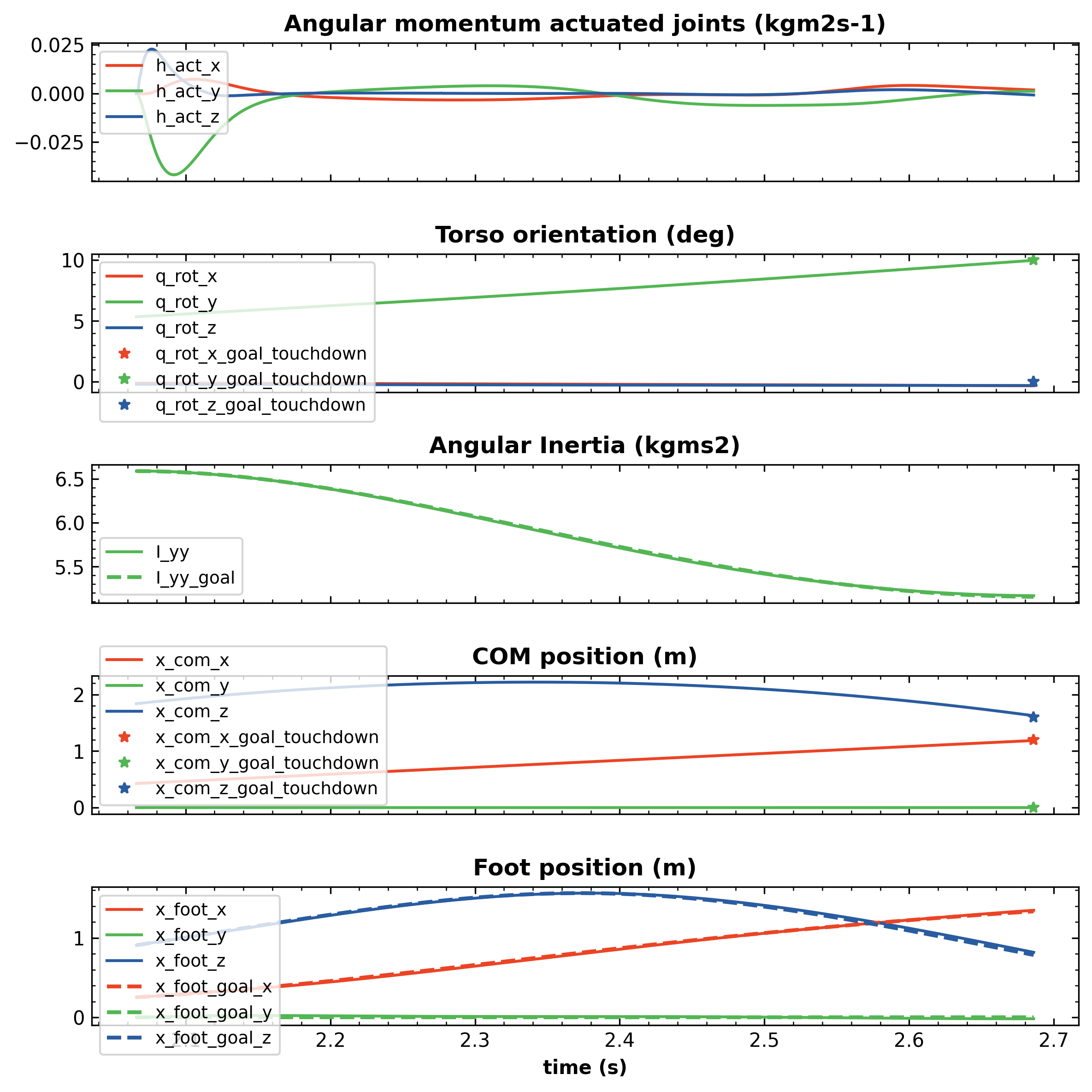}
  \caption{  \textbf{Flight Inertia Shaping.} Top: Actuated joints contribution to angular momentum. 2nd row: Torso orientation trajectory and goal. Center: Inertia, yy-component. 4th row: Center-of-mass trajectory and goal. Bottom row: Foot position.}
  \label{fig_lineplot_boxjump_up_inertia_shaping}
\end{figure}

The flight phase performance is displayed in Figure \ref{fig_lineplot_boxjump_up_inertia_shaping}. As indicated in section \ref{sec_planning_flight}, controlling the actuated joint's contributions to the total angular momentum, as well as the inertia tensor is essential in accomplishing the desired torso orientation at touchdown. The rotational inertia at the moment of takeoff is excessive, if uncontrolled this would lead to the robot landing at an under-rotated state. By smoothly shaping the inertia during flight the robot is able accelerate its rate of rotation and land at the desired orientation. The center-of-mass path cannot be influenced during flight, it is exclusively determined by the robot's state at take-off. During the flight phase, the landing foot's linear position and orientation are controlled in preparation for touchdown. 

\begin{figure}[H]
  \centering
  \includegraphics[width=.48\textwidth]{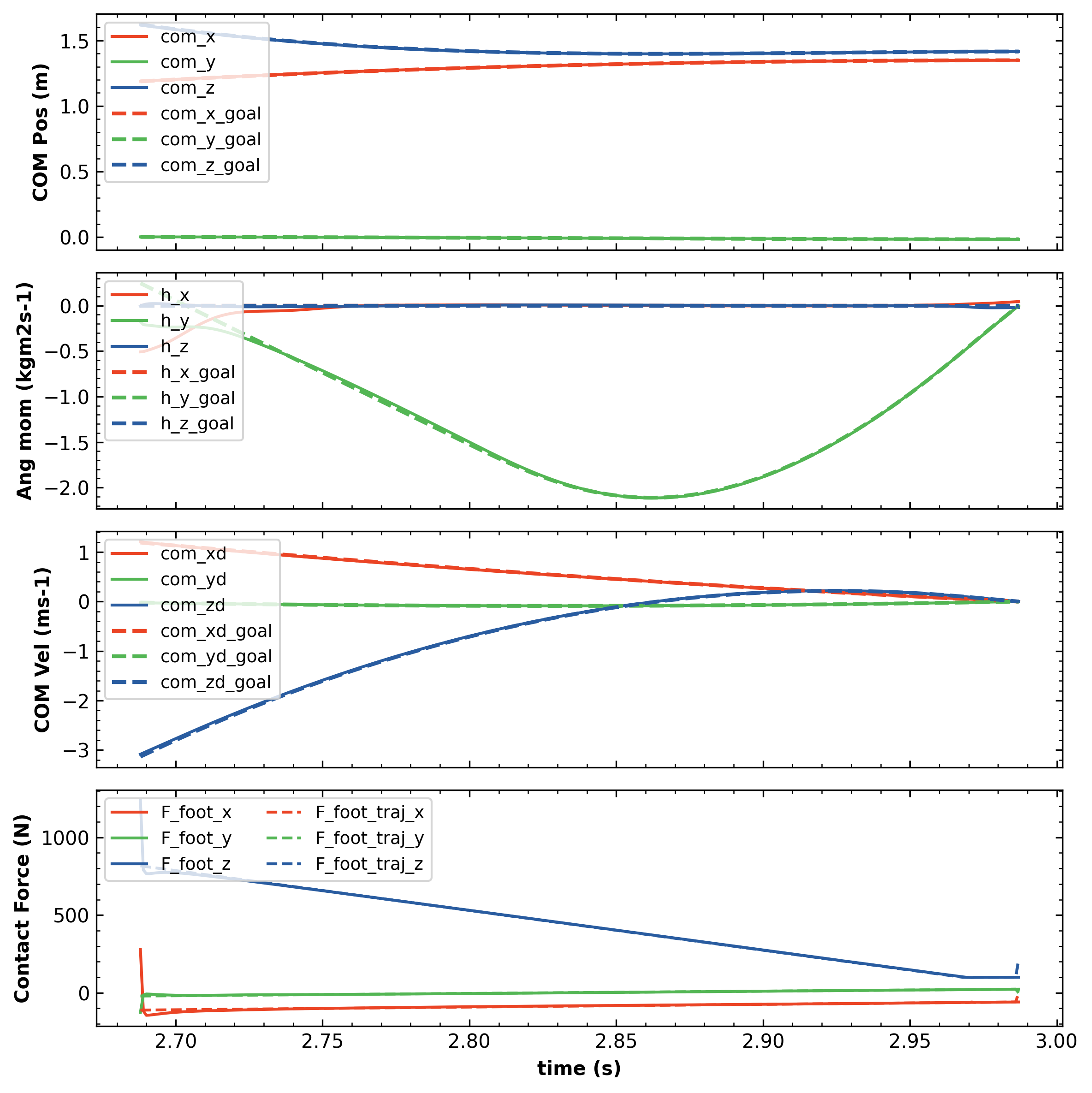}
  \caption{    \textbf{Landing Trajectory Tracking.}  Top, 3rd row: center-of-mass position and velocity, respectively. 2nd row: total angular momentum. Bottom: Resultant Foot contact force. }
  \label{fig_lineplot_boxjump_up_landing}
\end{figure}

\begin{figure}[H]
  \centering
  \includegraphics[width=.48\textwidth]{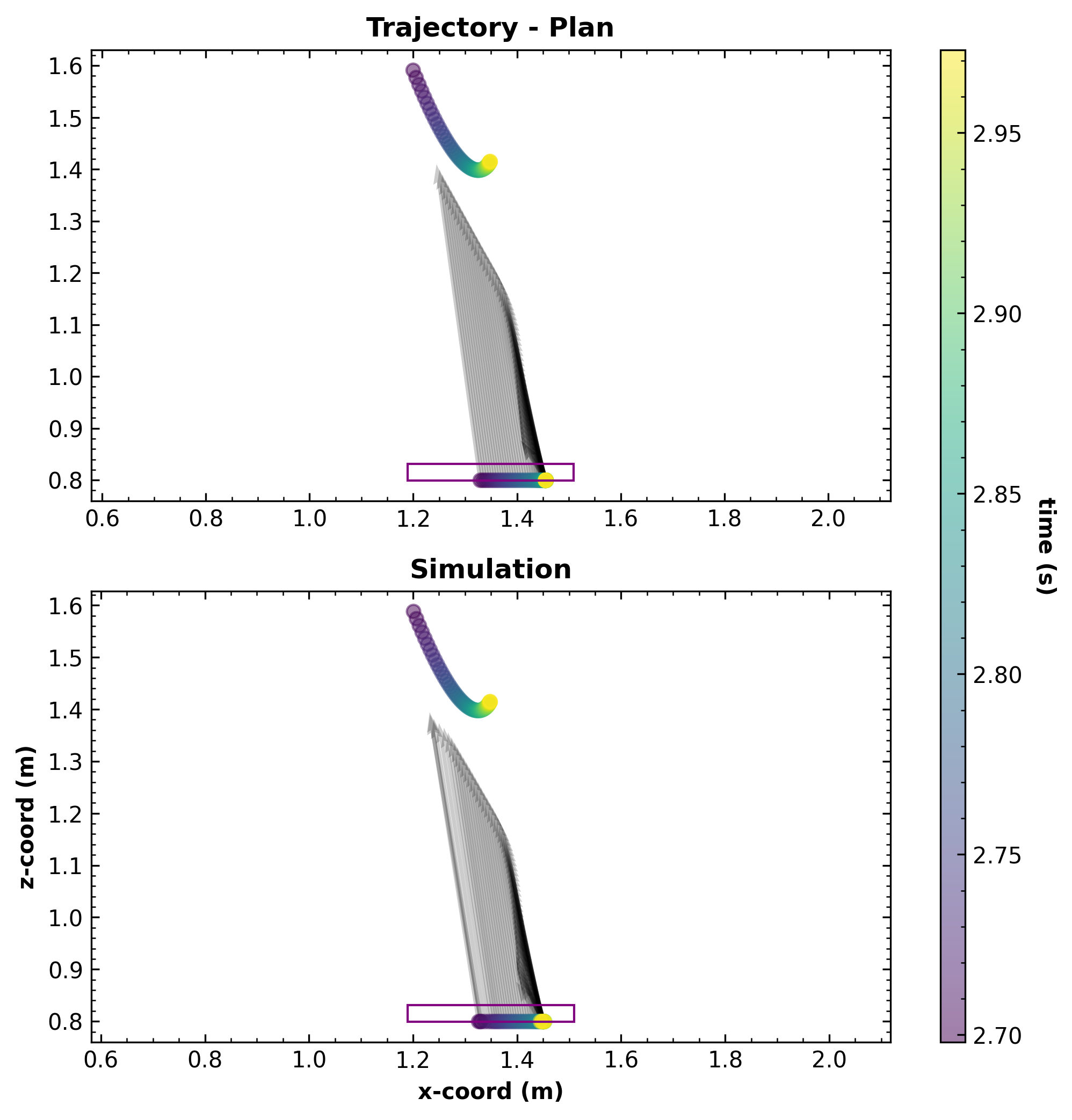}
  \caption{
  \textbf{Landing Scatter Plot.} 
  Top: Planned trajectories. Bottom: Measured simulation states. Black vectors represent resultant ground contact force. The colored point sequences represent center-of-mass (upper) and center-of-pressure (lower). Purple rectangle indicates the foot support. (Plotting every 5th data point.)}
  \label{fig_quiver_boxjump_up_landing}
\end{figure}

Figures \ref{fig_lineplot_boxjump_up_landing} and \ref{fig_quiver_boxjump_up_landing} illustrate the controller performance during the landing phase. The center-of-mass position converges to a resting position within the foot-support boundaries. The center-of-mass velocity as well as angular momentum are dissipated. The resultant contact force trajectory's direction, magnitude and center-of-pressure are well-aligned between plan and simulation.

\begin{figure}[H]
  \centering
  \includegraphics[width=.48\textwidth]{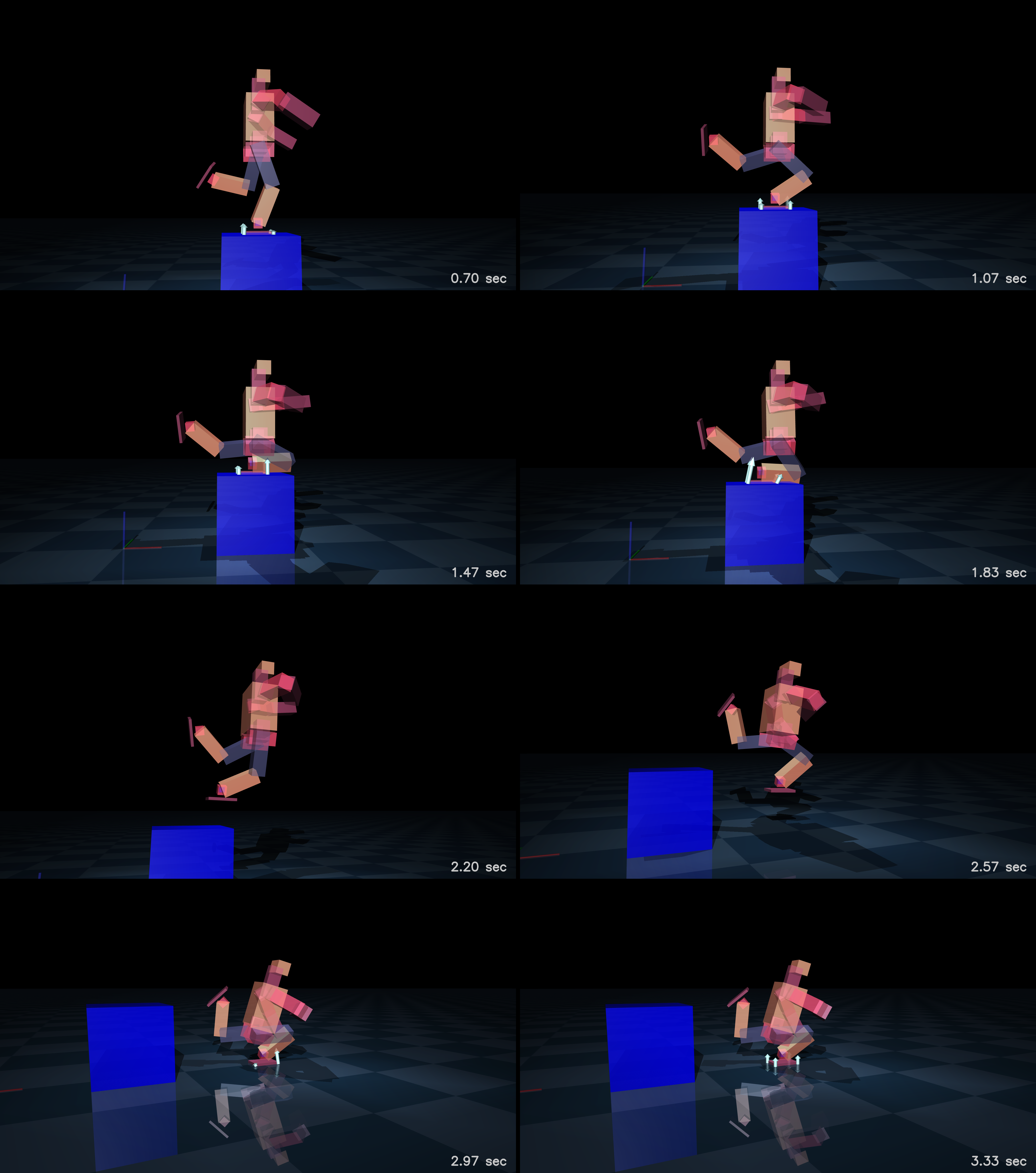}
  \caption{ \textbf{Box Jump Down Rendering.}  1st row: coiling phase, 2nd row: launch phase, 3rd row: flight phase, 4th row: landing phase. The 4 vectors indicate contact forces between foot and ground.}
  \label{fig_grid_boxjump_down}
\end{figure}

Figure \ref{fig_grid_boxjump_down} displays the downwards box jump utilizing the same trajectory optimization and controller as described above.

\subsection{Robustness Analysis}

\begin{figure}[H]
  \centering
  \includegraphics[width=.48\textwidth]{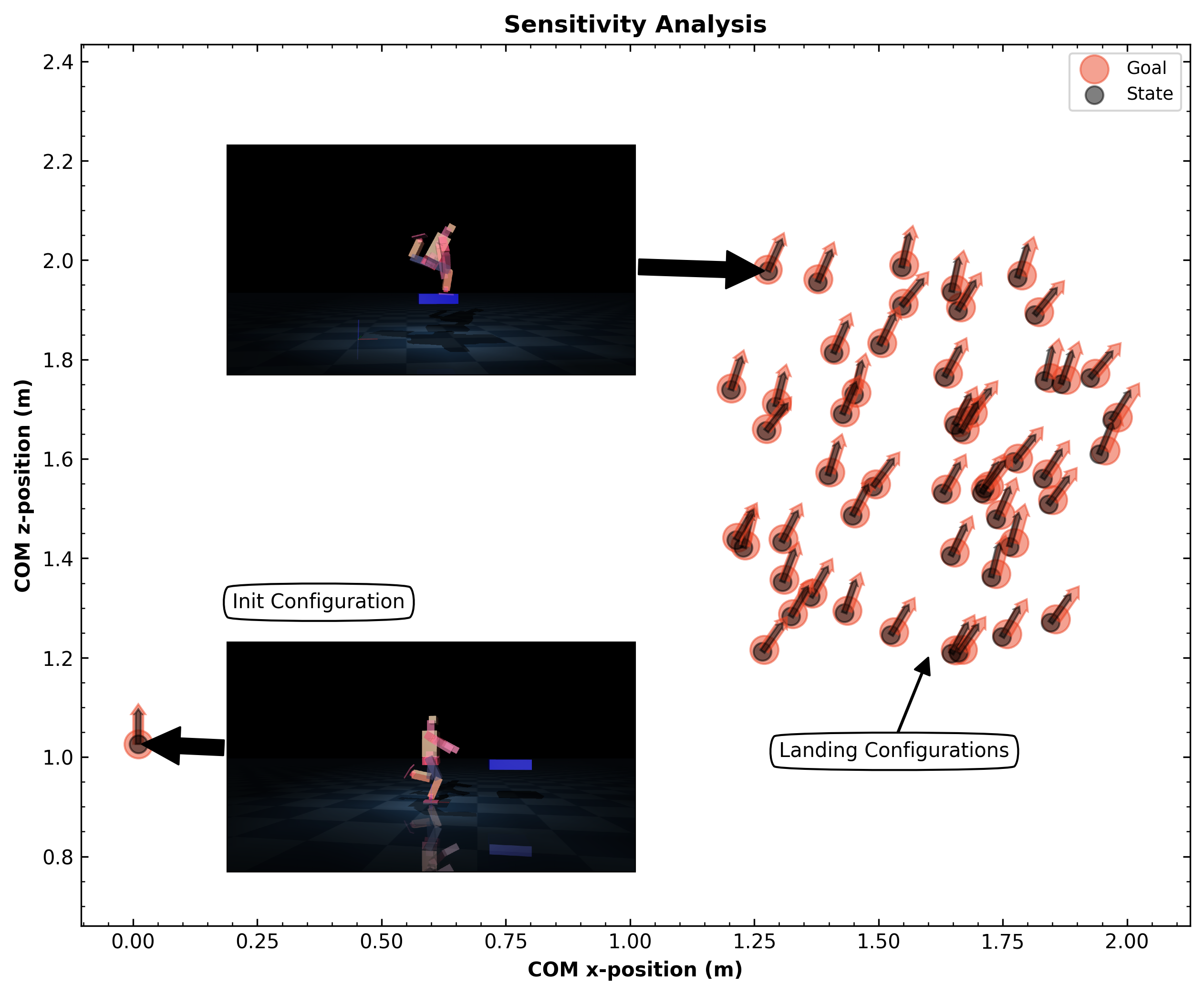}
  \caption{ 
    \textbf{Robustness Analysis.} 50 jumping simulations to arbitrary targets. Bottom-left: takeoff configuration. Right: 50 landing configurations. Circles indicate center-of-mass positions (State and Goal), arrows indicate torso orientation (up-vertical means upright torso). }
  \label{fig_sensitivity_boxjump}
\end{figure}

Establishing the controller's ability to accomplish arbitrary targets (within the bounds of feasibility), a variety of goal configurations are tested in simulation. Figure \ref{fig_sensitivity_boxjump} illustrates the results of 50 jumping simulations. A target center-of-mass position as well as a target torso orientation are randomly (uniformly) sampled within $\bm{x}_\mathrm{com}^\mathrm{touchdown} \in [1.2m, 2m]  $ and $\bm{q}_\mathrm{torso_y}^\mathrm{touchdown} \in [10^\circ, 40^\circ] $. The desired landing configurations are generally well-accomplished, however a slight bias of the center-of-mass position is evident, presumably due to unmodeled contact dynamics during the launch phase. Touchdown orientation is precisely accomplished, again, thanks to controllability during flight. Error statistics: $\varepsilon(\bm{x}_\mathrm{com}^\mathrm{touchdown}) = [\mu:0.888cm, \sigma:0.288cm]$. $\varepsilon(\bm{q}_\mathrm{torso_y}^\mathrm{touchdown}) = [\mu:0.094^\circ, \sigma:0.020^\circ]$.

\subsection{$360^\circ$ Front Flip and Constraints}

\begin{figure}[H]
  \centering
  \includegraphics[width=.48\textwidth]{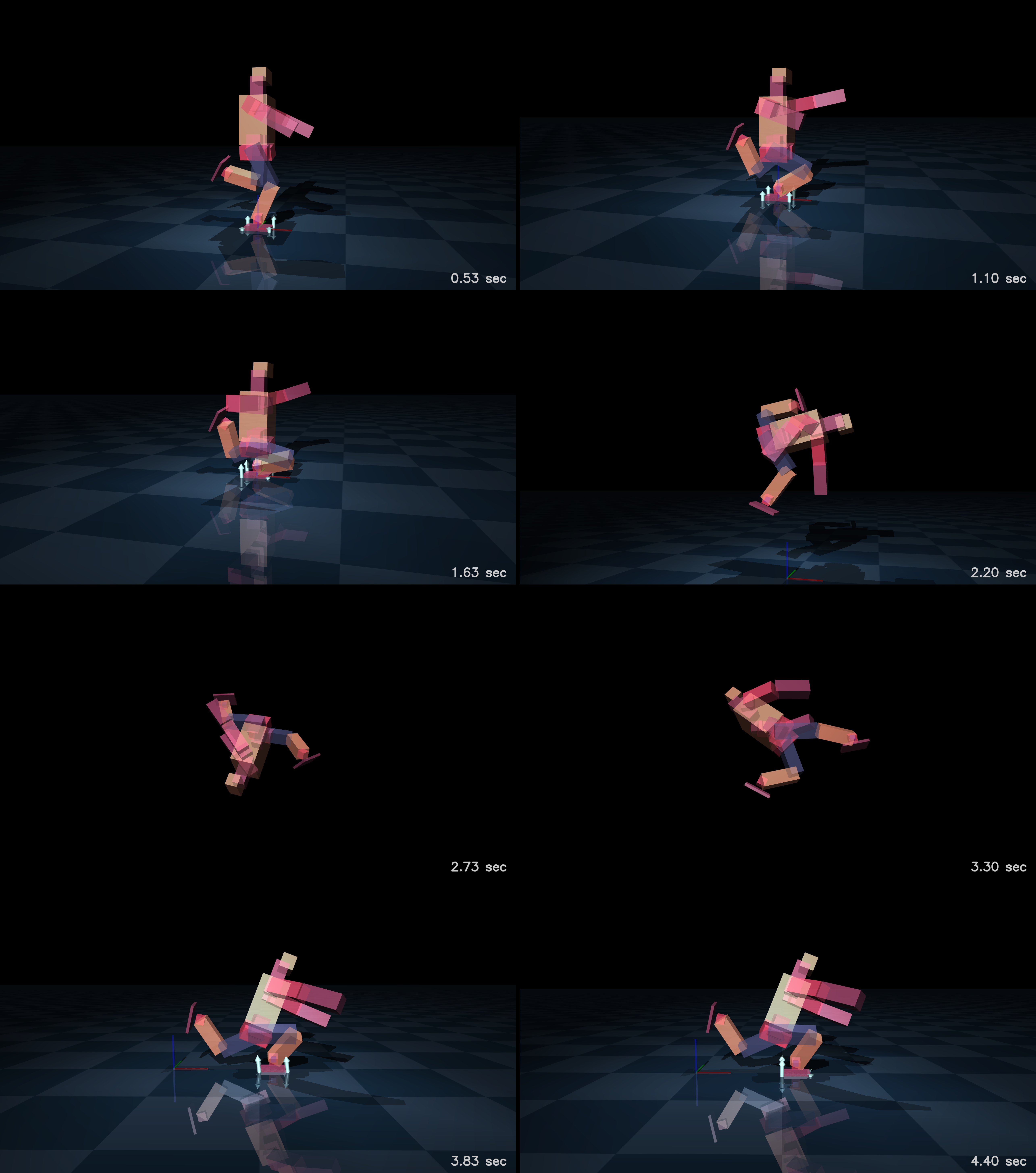}
  \caption{ 
    \textbf{Front Flip Rendering.} The robot performs a complete front flip.}
  \label{fig_grid_flip_forward}
\end{figure}

In this experiment, the robot performs a $360^\circ$ front flip (about the global y-coordinate). Again, the sequence entails launch-flight-land. Figure \ref{fig_grid_flip_forward} illustrates the entire sequence's rendering in simulation.

As established here and in the above experiments, the controller is able to track the robot's center-of-mass, inertias and momentums. However, because these trajectories are planned in the model abstraction, which is unaware of the robot's generalized coordinates, the whole-body controller needs to ensure that no constrains such as joint limits, self-collisions or undesired collisions with the environment occur. Utilizing the example of this front-flip experiment, these constraint tasks are described next.

\begin{figure}[H]
  \centering
  \includegraphics[width=.48\textwidth]{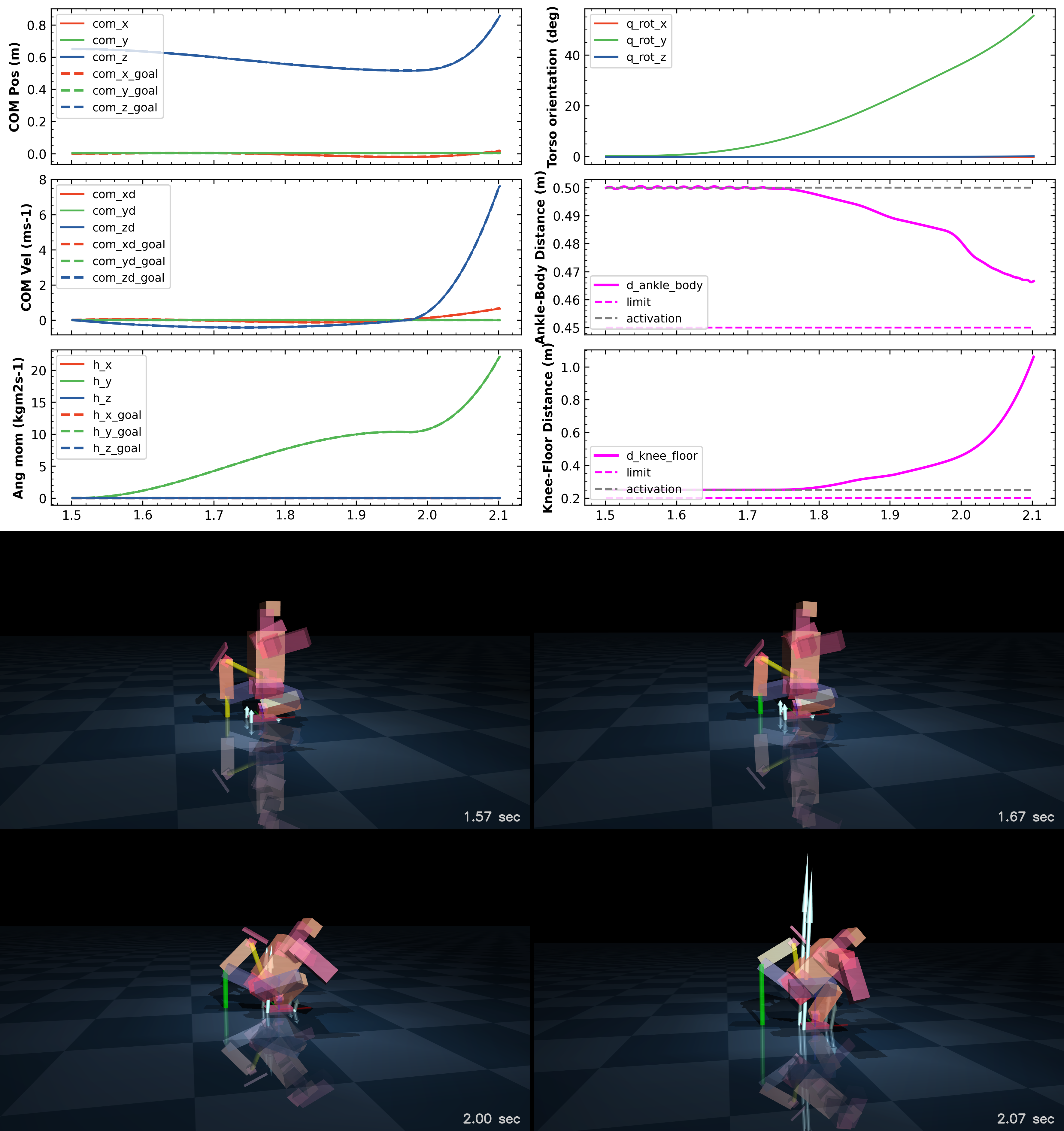}
  \caption{\textbf{Collision Avoidance during Launch.} Bottom: The green and yellow lines indicate shortest distances between two collision geometry pairs, ankle-torso and knee-floor. Yellow color indicates activated collision avoidance, green color indicates distance beyond activation. Top: Quantitative analysis. }
  \label{fig_collision_avoidance}
\end{figure}

Figure \ref{fig_collision_avoidance} displays collision avoidance (both self-collision and environmental) during the launch phase. Illustrating the point, two collision-pairs are displayed: a) ankle-torso, and b) knee-floor. 

As evident in the plots, controlling the trajectory-optimized quantities only would lead to collisions and constraint violations. The controller is able to avoid collisions while still tracking the desired trajectories without detriment. This is one of the circumstances where a larger number of degrees-of-freedom, i.e. a more complex robot model, is beneficial.

\begin{figure}[H]
  \centering
  \includegraphics[width=.48\textwidth]{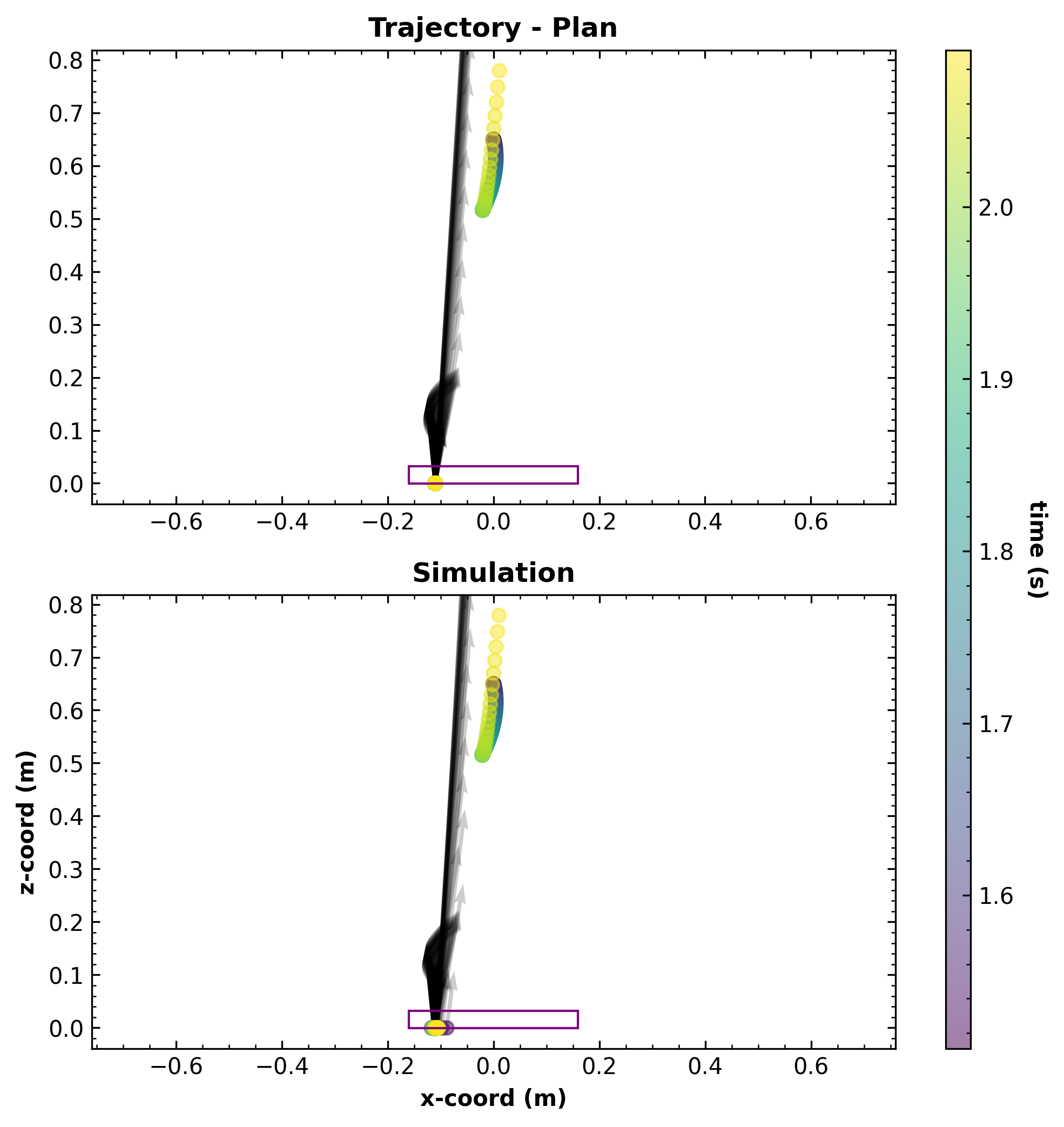}
  \caption{Figure \textbf{Front flip Launch Scatter Plot.} Top: Planned trajectories. Bottom: Measured simulation states. Black vectors represent resultant ground contact force. The colored point sequences represent center-of-mass (upper) and center-of-pressure (lower). Purple rectangle indicates the foot support. (Plotting every 5th data point.)}
  \label{fig_quiver_frontflip_launch}
\end{figure}

Figure \ref{fig_quiver_frontflip_launch} compares the associated launch com-path and contact forces between planned trajectory and during closed-loop execution. The optimizer initially lowers the com trajectory before accelerating upwards and forwards, together with scheduling an angular momentum ramp towards the end. This coordination allows the resultant contact forces to remain within the support zone. 

\begin{figure}[H]
  \centering
  \includegraphics[width=.48\textwidth]{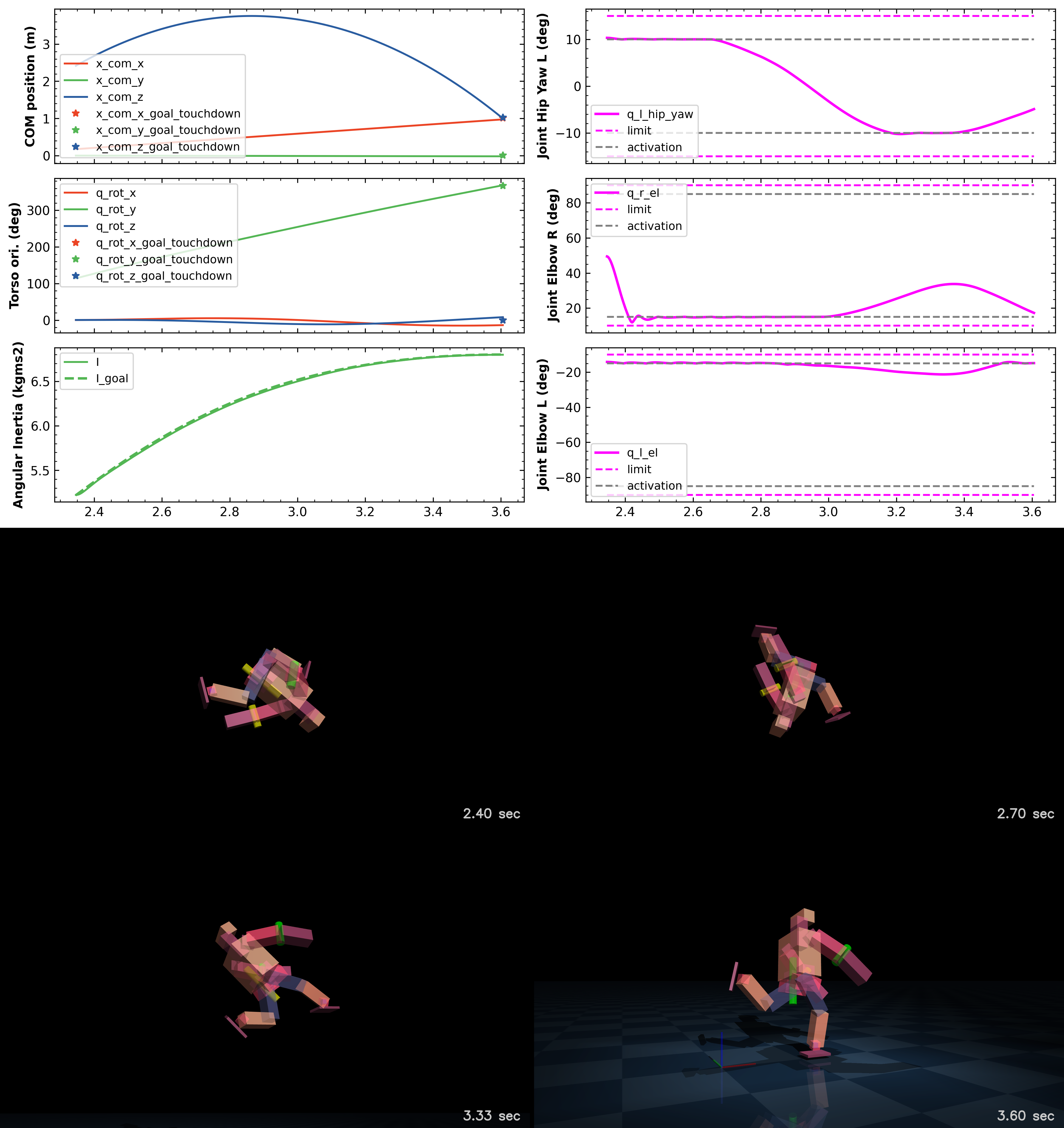}
  \caption{ \textbf{Inertia Shaping under Constraint Avoidance.} Bottom: The green and yellow cylinders indicate joint limit task states of three joints: Right and left elbows, and left hip-yaw. Green color indicates inactive constraint task (i.e. beyond activation margin), yellow color indicates activated joint limit avoidance. Centroidal quantities and inertias are precisely tracked  while enforcing the constraints.}
  \label{fig_jointlimit_avoidance}
\end{figure}

Figure \ref{fig_jointlimit_avoidance} exemplifies joint-limit avoidance during the flight phase. While joint limits apply to all joints, only the state of three are rendered here for the purpose of clarity:  the robot's left and right elbows and left hip-yaw joint. All three joint limits would ordinarily have been exceeded. The joint limit task activates at a margin of $5^\circ$ before the upper and lower hard limits. Due to the sufficiently high number of degrees-of-freedom, the controller manages to track the trajectories including inertia-shaping with precision while avoiding the limits. 

Notably, the robot launches with excess angular momentum, leading the flight controller to elevate the system's inertia (hence the robot's spreading limbs) in order to slow down angular velocity and land at the desired orientation.

\subsection{$\mathrm{180^\circ}$ Twisting Jump - Posture Control}

\begin{figure}[H]
  \centering
  \includegraphics[width=.48\textwidth]{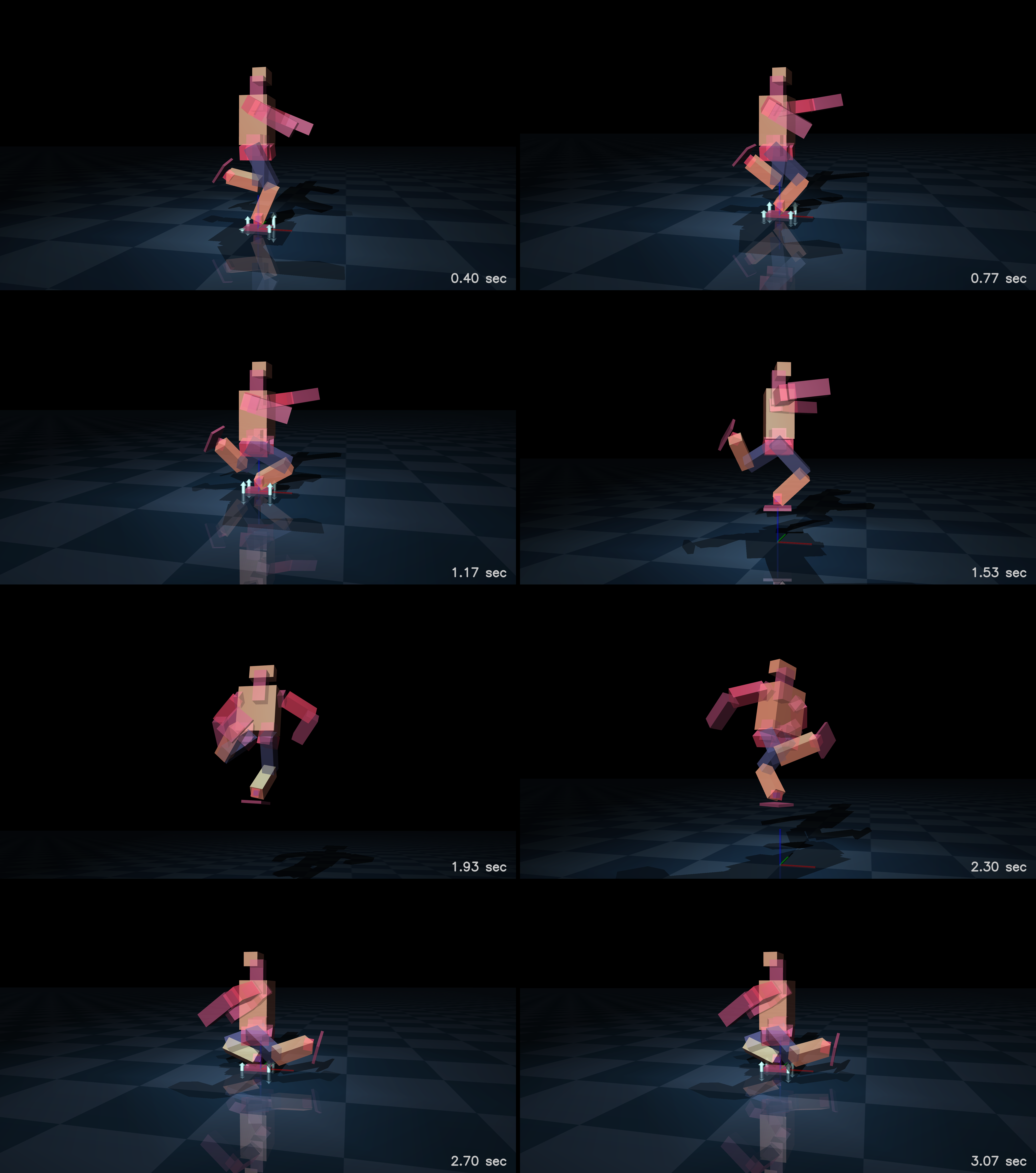}
  \caption{ \textbf{Twisting Jump Rendering.} The robot performs a $180^\circ$ twisting jump, ensuring the torso terminates at a desired orientation (upright and axis-aligned) via posture control.}
  \label{fig_grid_twist}
\end{figure}

The previous experiments illustrated a variety of individual dynamic maneuvers. In order to prepare the robot for sequences of compounded maneuvers and to enable stable limit-cycles, the robot must be able to terminate at desire postures. 
This ability is illustrated in this experiment, where the robot performs a $\mathrm{180^\circ}$ twisting jump, rotating about the global z-coordinate. Figure \ref{fig_grid_twist} displays the entire sequence. 

\begin{figure}[H]
  \centering
  \includegraphics[width=.48\textwidth]{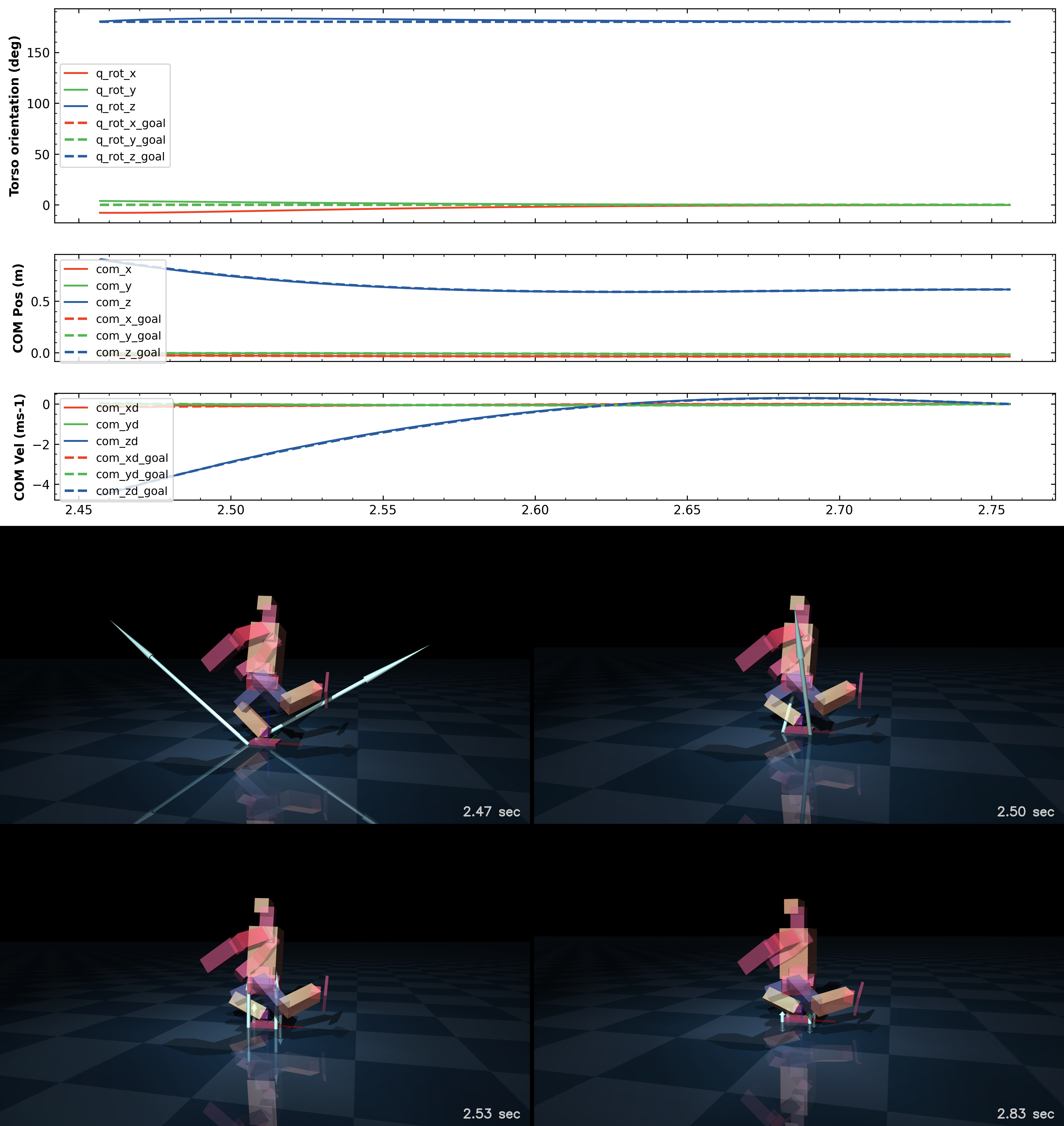}
  \caption{ \textbf{Posture control while landing.}  Bottom, chronologically: Touchdown with angled torso, linear and angular momentums dissipated via foot support, posture controller transitions the torso into the desired upright and axis-aligned state at termination.}
  \label{fig_twist_posture_land}
\end{figure}

The robot builds angular momentum about the z-axis during launch, note the generated moment at the foot contact, shapes inertia as to land after $180^\circ$ rotation and dissipates into a steady state. 
The posture goal in this example is for the robot's torso to terminate at a desired orientation (upright and axis-aligned with the global coordinate system). Figure \ref{fig_twist_posture_land} analyzes the robot's behavior during the landing phase. The robot's linear and angular momentums are dissipated as planned by the trajectory optimizer. At the moment of touchdown, the torso is angled which the posture controller transitions into the desired upright and axis-aligned final state.

\section{Conclusion}

This paper presented an architecture combining trajectory optimization and a whole-body controller, that through a model abstraction, enables humanoid robots to perform acrobatic motions. Importantly, no approximations of the articulated rigid body dynamics model was needed and the optimizer as well as the controller operated with awareness of the complete underlying equations-of-motion. 
Simulations and analysis of recorded data demonstrated the system's effectiveness on various highly dynamic motions including box jumps, flips and twisting jumps. 

The controller could be improved by modeling the contact dynamics between foot and ground. Since this is a model-based approach, this architecture generalizes to various robot types. Additional robots with distinct kinematics and mass properties could be loaded to tested to demonstrate this capability. 

The robot in the above simulations used a single support surface in order to demonstrate the most-challenging condition with the smallest contact support area. The architecture generalizes to multiple support surfaces (e.g. launching and landing on both feet) with minor modifications. 

An immediate extension would be the utilization of the trajectory optimizer in an MPC type of architecture for additional robustness. Given the simplicity of the transcribed optimization programs, there should be no theoretical obstacles to this.

\normalsize
\bibliography{references}

\begin{thebibliography}{26}
\providecommand{\natexlab}[1]{#1}
\providecommand{\url}[1]{\texttt{#1}}
\expandafter\ifx\csname urlstyle\endcsname\relax
  \providecommand{\doi}[1]{doi: #1}\else
  \providecommand{\doi}{doi: \begingroup \urlstyle{rm}\Url}\fi

\bibitem[Vukobratovi{\'c} and Borovac(2004)]{vukobratovic2004zero}
Miomir Vukobratovi{\'c} and Branislav Borovac.
\newblock Zero-moment point—thirty five years of its life.
\newblock \emph{International journal of humanoid robotics}, 1\penalty0 (01):\penalty0 157--173, 2004.

\bibitem[Sardain and Bessonnet(2004)]{sardain2004forces}
Philippe Sardain and Guy Bessonnet.
\newblock Forces acting on a biped robot. center of pressure-zero moment point.
\newblock \emph{IEEE Transactions on Systems, Man, and Cybernetics-Part A: Systems and Humans}, 34\penalty0 (5):\penalty0 630--637, 2004.

\bibitem[Goswami(1999)]{goswami1999postural}
Ambarish Goswami.
\newblock Postural stability of biped robots and the foot-rotation indicator (fri) point.
\newblock \emph{The International Journal of Robotics Research}, 18\penalty0 (6):\penalty0 523--533, 1999.

\bibitem[Spong and Bhatia(2003)]{spong2003further}
Mark~W Spong and Gagandeep Bhatia.
\newblock Further results on control of the compass gait biped.
\newblock In \emph{Proceedings 2003 IEEE/RSJ International Conference on Intelligent Robots and Systems (IROS 2003)(Cat. No. 03CH37453)}, volume~2, pages 1933--1938. IEEE, 2003.

\bibitem[Blickhan and Full(1993)]{blickhan1993similarity}
Reinhard Blickhan and RJ~Full.
\newblock Similarity in multilegged locomotion: bouncing like a monopode.
\newblock \emph{Journal of Comparative Physiology A}, 173:\penalty0 509--517, 1993.

\bibitem[Goswami and Kallem(2004)]{goswami2004rate}
Ambarish Goswami and Vinutha Kallem.
\newblock Rate of change of angular momentum and balance maintenance of biped robots.
\newblock In \emph{IEEE International Conference on Robotics and Automation, 2004. Proceedings. ICRA'04. 2004}, volume~4, pages 3785--3790. IEEE, 2004.

\bibitem[Sano and Furusho(1990)]{sano1990realization}
Akihito Sano and Junji Furusho.
\newblock Realization of natural dynamic walking using the angular momentum information.
\newblock In \emph{Proceedings., IEEE International Conference on Robotics and Automation}, pages 1476--1481. IEEE, 1990.

\bibitem[Raibert(1984)]{raibert1984hopping}
Marc~H Raibert.
\newblock Hopping in legged systems—modeling and simulation for the two-dimensional one-legged case.
\newblock \emph{IEEE Transactions on Systems, Man, and Cybernetics}, \penalty0 (3):\penalty0 451--463, 1984.

\bibitem[Raibert(1986)]{raibert1986legged}
Marc~H Raibert.
\newblock \emph{Legged robots that balance}.
\newblock MIT press, 1986.

\bibitem[Guizzo(2019)]{guizzo2019leaps}
Erico Guizzo.
\newblock By leaps and bounds: An exclusive look at how boston dynamics is redefining robot agility.
\newblock \emph{IEEE Spectrum}, 56\penalty0 (12):\penalty0 34--39, 2019.

\bibitem[Qi et~al.(2023)Qi, Chen, Yu, Huang, Liu, Meng, and Huang]{qi2023vertical}
Haoxiang Qi, Xuechao Chen, Zhangguo Yu, Gao Huang, Yaliang Liu, Libo Meng, and Qiang Huang.
\newblock Vertical jump of a humanoid robot with cop-guided angular momentum control and impact absorption.
\newblock \emph{IEEE Transactions on Robotics}, 39\penalty0 (4):\penalty0 3154--3166, 2023.

\bibitem[Dai et~al.(2014)Dai, Valenzuela, and Tedrake]{dai2014whole}
Hongkai Dai, Andr{\'e}s Valenzuela, and Russ Tedrake.
\newblock Whole-body motion planning with centroidal dynamics and full kinematics.
\newblock In \emph{2014 IEEE-RAS International Conference on Humanoid Robots}, pages 295--302. IEEE, 2014.

\bibitem[He et~al.(2024)He, Wu, Zhang, Zhang, Shi, Liu, Sun, Su, and Leng]{he2024cdm}
Zhicheng He, Jiayang Wu, Jingwen Zhang, Shibowen Zhang, Yapeng Shi, Hangxin Liu, Lining Sun, Yao Su, and Xiaokun Leng.
\newblock Cdm-mpc: An integrated dynamic planning and control framework for bipedal robots jumping.
\newblock \emph{IEEE Robotics and Automation Letters}, 2024.

\bibitem[Wang et~al.(2023)Wang, Xin, Xin, Mistry, Vijayakumar, and Kormushev]{wang2023unified}
Ke~Wang, Guiyang Xin, Songyan Xin, Michael Mistry, Sethu Vijayakumar, and Petar Kormushev.
\newblock A unified model with inertia shaping for highly dynamic jumps of legged robots.
\newblock \emph{Mechatronics}, 95:\penalty0 103040, 2023.

\bibitem[Bang et~al.(2024)Bang, Lee, Gonzalez, and Sentis]{bang2024variable}
Seung~Hyeon Bang, Jaemin Lee, Carlos Gonzalez, and Luis Sentis.
\newblock Variable inertia model predictive control for fast bipedal maneuvers.
\newblock \emph{arXiv preprint arXiv:2407.16811}, 2024.

\bibitem[Khatib et~al.(2008)Khatib, Sentis, and Park]{khatib2008unified}
Oussama Khatib, Luis Sentis, and Jae-Heung Park.
\newblock A unified framework for whole-body humanoid robot control with multiple constraints and contacts.
\newblock In \emph{European Robotics Symposium 2008}, pages 303--312. Springer, 2008.

\bibitem[Khatib(1990)]{khatib1990motion}
Oussama Khatib.
\newblock Motion/force redundancy of manipulators.
\newblock In \emph{Proceedings of Japan-USA Symposium on Flexible Automation}, volume~1, pages 337--342, 1990.

\bibitem[Khatib et~al.(2004)Khatib, Sentis, Park, and Warren]{khatib2004whole}
Oussama Khatib, Luis Sentis, Jaeheung Park, and James Warren.
\newblock Whole-body dynamic behavior and control of human-like robots.
\newblock \emph{International Journal of Humanoid Robotics}, 1\penalty0 (01):\penalty0 29--43, 2004.

\bibitem[Khatib et~al.(2022)Khatib, Jorda, Park, Sentis, and Chung]{khatib2022constraint}
Oussama Khatib, Mikael Jorda, Jaeheung Park, Luis Sentis, and Shu-Yun Chung.
\newblock Constraint-consistent task-oriented whole-body robot formulation: Task, posture, constraints, multiple contacts, and balance.
\newblock \emph{The International Journal of Robotics Research}, 41\penalty0 (13-14):\penalty0 1079--1098, 2022.

\bibitem[Orin et~al.(2013)Orin, Goswami, and Lee]{orin2013centroidal}
David~E Orin, Ambarish Goswami, and Sung-Hee Lee.
\newblock Centroidal dynamics of a humanoid robot.
\newblock \emph{Autonomous robots}, 35:\penalty0 161--176, 2013.

\bibitem[Ess{\'e}n(1993)]{essen1993average}
Hanno Ess{\'e}n.
\newblock Average angular velocity.
\newblock \emph{European journal of physics}, 14\penalty0 (5):\penalty0 201, 1993.

\bibitem[Boyle(2017)]{boyle2017integration}
Michael Boyle.
\newblock The integration of angular velocity.
\newblock \emph{Advances in Applied Clifford Algebras}, 27\penalty0 (3):\penalty0 2345--2374, 2017.

\bibitem[Khatib(1986)]{khatib1986real}
Oussama Khatib.
\newblock Real-time obstacle avoidance for manipulators and mobile robots.
\newblock \emph{The international journal of robotics research}, 5\penalty0 (1):\penalty0 90--98, 1986.

\bibitem[Ericson(2004)]{ericson2004real}
Christer Ericson.
\newblock \emph{Real-time collision detection}.
\newblock Crc Press, 2004.

\bibitem[Todorov et~al.(2012)Todorov, Erez, and Tassa]{todorov2012mujoco}
Emanuel Todorov, Tom Erez, and Yuval Tassa.
\newblock Mujoco: A physics engine for model-based control.
\newblock In \emph{2012 IEEE/RSJ International Conference on Intelligent Robots and Systems}, pages 5026--5033. IEEE, 2012.
\newblock \doi{10.1109/IROS.2012.6386109}.

\bibitem[W{\"a}chter and Biegler(2006)]{wachter2006implementation}
Andreas W{\"a}chter and Lorenz~T Biegler.
\newblock On the implementation of an interior-point filter line-search algorithm for large-scale nonlinear programming.
\newblock \emph{Mathematical programming}, 106:\penalty0 25--57, 2006.

\end{thebibliography}

\end{document}